\documentclass{article}

\usepackage{microtype}
\usepackage{graphicx}
\usepackage{subcaption}
\usepackage{booktabs} 

\usepackage{hyperref}

\usepackage[preprint]{icml2026}

\usepackage{amsmath}
\usepackage{amssymb}
\usepackage{mathtools}
\usepackage{amsthm}

\usepackage[capitalize,noabbrev]{cleveref}

\theoremstyle{plain}

\theoremstyle{definition}

\theoremstyle{remark}

\usepackage[textsize=tiny]{todonotes}

\usepackage{graphicx}
\usepackage{wrapfig}
\usepackage{multirow} 
\usepackage{makecell}
\usepackage{bbding}
\usepackage{pifont}
\usepackage{colortbl}
\usepackage{hyperref}

\newcommand{\emoji}[2][1.2em]{\raisebox{-0.2\height}{\includegraphics[height=#1]{#2}}}
\definecolor{fbApp}{HTML}{ffe4e3}
\newcommand{\rowc}{\rowcolor{fbApp}}

\icmltitlerunning{Empowering Time Series Analysis with Large-Scale Multimodal Pretraining}

\begin{document}

\twocolumn[
  \icmltitle{Empowering Time Series Analysis with Large-Scale Multimodal Pretraining}



  \icmlsetsymbol{equal}{*}

  \begin{icmlauthorlist}
    \icmlauthor{Peng Chen}{yyy}
    \icmlauthor{Siyuan Wang}{yyy}
    \icmlauthor{Shiyan Hu}{yyy}
    \icmlauthor{Xingjian Wu}{yyy}
    \icmlauthor{Yang Shu}{yyy}
    \icmlauthor{Zhongwen Rao}{comp}
    \icmlauthor{Meng Wang}{comp}
    \icmlauthor{Yijie Li}{comp}
    \icmlauthor{Bin Yang}{yyy}
    \icmlauthor{Chenjuan Guo}{yyy}
  \end{icmlauthorlist}

  \icmlaffiliation{yyy}{East China Normal University, Shanghai, China}
  \icmlaffiliation{comp}{HuaWei, ShenZhen, China}

  \icmlcorrespondingauthor{Chenjuan Guo}{cjguo@dase.ecnu.edu.cn}
  \icmlkeywords{Machine Learning, ICML}

  \vskip 0.3in
]



\printAffiliationsAndNotice{}  

\begin{abstract}
While existing time series foundation models primarily rely on large-scale unimodal pretraining, they lack complementary modalities to enhance time series understanding. Building multimodal foundation models is a natural next step, but it faces key challenges: 
1) lack of a unified multimodal pretraining paradigm and large-scale multimodal corpora for time series analysis;
2) how to effectively integrate heterogeneous modalities and enhance model generalization. To address these challenges, we take an early step toward multimodal foundation models for time series analysis. 
We first propose a multimodal pretraining paradigm that leverages time series with endogenous modalities (derived images and text) and exogenous knowledge (real-world news), providing a comprehensive multi-view perspective for time series analysis. To support this, we develop an automated data construction pipeline to curate MM-TS, the first large-scale multimodal time series dataset spanning six domains, with up to one billion points. Then we propose HORAI, a frequency-enhanced multimodal foundation model.
It integrates two core components: the Frequency-enhanced Cross-Modality Encoder and the Time-Frequency Decoder, designed to effectively fuse multimodal features and enhance model generalization across modalities and domains. After pretraining on MM-TS, HORAI achieves state-of-the-art zero-shot performance on time series forecasting and anomaly detection tasks, demonstrating strong generalization.
\end{abstract}

\section{Introduction}

Time series analysis is widely applied across diverse domains, including energy management, medical monitoring, and financial forecasting. Existing time series analysis approaches, ranging from time-series-specific models \citep{dlinear, patchtst, itransformer, pathformer} to recent time series foundation models \citep{moirai, units, timemoe, flame}, primarily rely on time series numerical modality to capture temporal patterns and uncover underlying regularities. While these methods have achieved competitive performance, this single-modality paradigm remains limited in its ability to capture the complex and multifaceted nature of real-world temporal dynamics \citep{intervention}.

At the same time, foundation models in NLP and multimodal learning \citep{gpt3,qwen, next-gpt,internvl} have shown that large-scale pretraining on massive datasets with complementary modalities can enhance generalization and adaptability across tasks. Inspired by these, we propose developing multimodal foundation models for time series analysis.
By incorporating additional modalities for pretraining, such as text and images, the model leverages textual semantics and visual information to better capture complex temporal dynamics and strengthen time series understanding.

\begin{figure*}[t]
    \centering  
\includegraphics[width=1\linewidth]{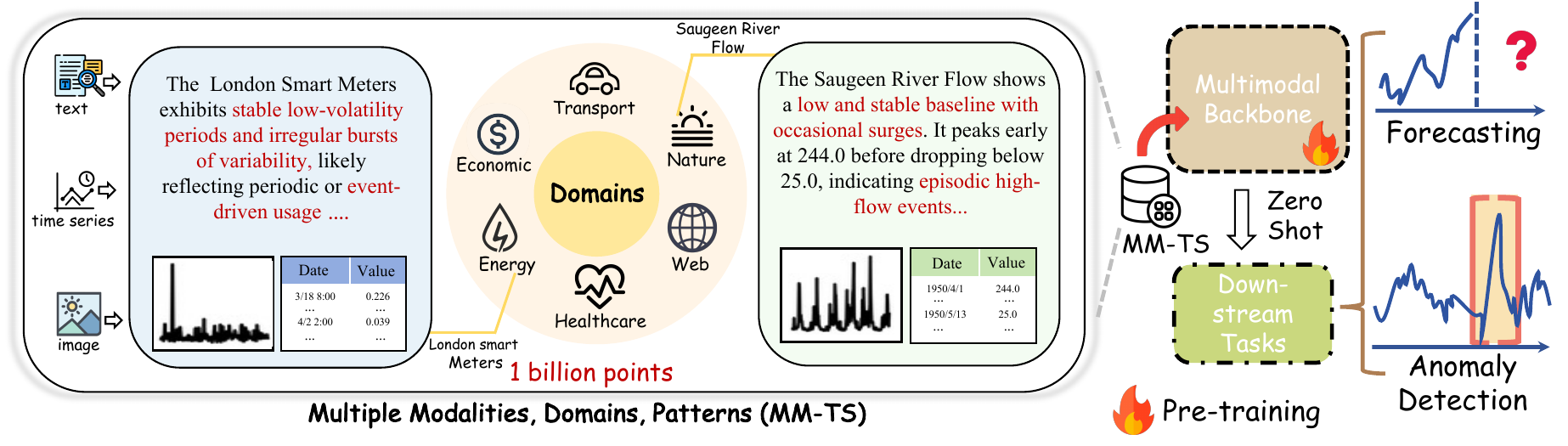}
\vspace{-2em}
\caption{Left: The large-scale multimodal time series dataset MM-TS is characterized by its coverage of various modalities, heterogeneous domains, and diverse temporal patterns. Right: The multimodal foundation model HORAI is pre-trained on the MM-TS dataset and evaluated on downstream scenarios and tasks.  }
\label{figs:dataset}
\vspace{-1em}
\end{figure*}

However, the development of multimodal foundation models faces several significant challenges. First, the field lacks a multimodal pretraining paradigm and large-scale multimodal corpora for time series analysis. 
Existing multimodal methods are tailored to specific datasets, lacking a unified pretraining paradigm for broad cross-modal analysis \cite{gpt4mts, tats}. 
Moreover, existing large-scale datasets contain only numerical time series and lack complementary modalities such as text or images, which limits their suitability for multimodal pretraining \cite{timer, moirai}.
Second, the architectural design for integrating different modalities in time series analysis remains underexplored. Each modality exhibits unique characteristics: text provides rich semantic information and offers a holistic, global description of events, whereas images capture localized details and spatial structures \citep{timevlm}. Directly fusing time series with textual or visual modalities \citep{distillation, distillation1} may result in suboptimal alignment and ineffective representation learning. Therefore, it is critical to design fusion mechanisms that explicitly leverage the unique characteristics of each modality.
Third, time series data from different domains exhibit diverse patterns, and the incorporation of multiple modalities further amplifies this diversity. Effectively modeling the heterogeneous patterns across modalities and domains, while enhancing the generalization ability of pretrained models, remains a challenge.
Consequently, advancing multi-modal foundation models for time series analysis requires further research and exploration.

In this paper, we take an early step toward developing multimodal foundation models for time series analysis. \textit{On the pretraining paradigm and dataset side}, we propose a multimodal pretraining paradigm that leverages time series with endogenous modalities (derived images and text) and exogenous knowledge (real-world news), providing a comprehensive multi-view perspective for time series analysis. Endogenous modalities capture intrinsic time series characteristics through semantic and visual structures, while exogenous news provides supplementary background context.
To support this, we develop an \textit{automated data construction pipeline} that maps raw sequences to visual and textual modalities, aligns them with external news sources, and incorporates multiple LLMs for rigorous quality verification.
This process curates MM-TS, the first large-scale multimodal time series dataset. 
As illustrated in Figure \ref{figs:dataset}, MM-TS integrates three modalities, spanning six diverse domains and a wide range of temporal patterns, with up to one billion points. The three modalities exhibit strong correlations and complementary characteristics, making MM-TS well-suited for multimodal pretraining to learn generalized representations. 

\textit{On the modeling side}, we propose \textbf{HORAI}, a frequency-enhanced multimodal time series foundation model built on an autoregressive architecture, which consists of two core components: Frequency-enhanced Cross-Modality Encoder and Time-Frequency Decoder. \textit{In the Frequency-enhanced Cross-Modality Encoder}, we leverage the correspondence between modality-specific information and different frequency components of time series to align multiple modalities and enhance time series understanding. Specifically, time series are decomposed into multiple frequency bands, where low-frequency components capture long-term dynamics and align with the global semantics embedded in text, while mid- and high-frequency components encode rapid variations that tend to correspond to the localized patterns present in visual inputs. 
\textit{In the Time-Frequency Decoder}, we design a Time-Frequency MoE-FFN to learn generalized multimodal representations from multi-domain data. We introduce a time-frequency router that dynamically assigns each token to the suitable expert based on both temporal and frequency features. 
By incorporating frequency-domain features, the router gains additional cues to better distinguish similar patterns and group them coherently, which enhances feature consistency and improves generalization across domains. 
Specifically, our contributions are as follows:
\begin{itemize}
    \item 
    \textbf{New Paradigm and Pretraining Datasets:} We propose a multimodal pretraining paradigm that leverages time series along with endogenous modalities (derived images and text) and exogenous knowledge (news), enhancing time series analysis from a comprehensive multi-view perspective. To support this, we develop an automated data construction pipeline to curate MM-TS, the first large-scale multimodal time series dataset. 

    \item
    \textbf{New Model:} We propose HORAI, a frequency-enhanced multimodal foundation model for Time series analysis, which incorporates two core components, the frequency-guided cross-modality encoder and the time-frequency decoder, designed to effectively fuse multimodal features and enhance model generalization across modalities and domains.

    \item
    \textbf{Comprehensive  Evaluation:} After pre-training on large-scale multimodal time series data, HORAI achieves state-of-the-art performance in time series forecasting and anomaly detection across zero-shot inference and few-shot learning situations, which demonstrates strong task versatility and generalization ability.

\end{itemize}

 

\vspace{-2em}
\section{Related work}

\subsection{Time Series Analysis}
Time series analysis spans a wide range of tasks, including forecasting and anomaly detection ~\citep{tfb, anomly_survey, tsb}. Existing approaches can be broadly divided into unimodal and multimodal methods. \textbf{Unimodal methods} focus on time series data and employ diverse architectures to model temporal dynamics and channel correlations. These include MLP-based models~\citep{dlinear, fits}, RNN-based models~\citep{deepar}, CNN-based models~\citep{Timesnet, moderntcn}, GNN-based models~\citep{zhao2023multiple, autocts}, as well as Transformer-based architectures for capturing long-range dependencies~\citep{crossformer, patchtst, pathformer,dtaf}. In contrast, \textbf{multimodal methods} integrate additional modalities or external knowledge to enhance time series analysis. One line of work introduces endogenous prompts, such as statistical information, channel semantics, or task-related descriptions, to enrich temporal representations~\citep{timllm, cctime, timevlm}. Another line of work leverages exogenous textual or visual modalities to provide additional contextual knowledge~\citep{tats, gpt4mts, chattime, timemmd}. Although these methods achieve competitive performance, most require retraining and extensive parameter tuning for each dataset, lacking zero-shot inference capabilities. While ChatTime~\citep{chattime} enables direct zero-shot inference, it suffers from precision loss due to data discretization and lacks rich multimodal characterizations. Further comparisons with ChatTime are in Appendix \ref{discussion chattime}.

\subsection{Time Series Foundation Models}
Recently, time series foundation models (TSFMs) have attracted increasing attention~\citep{chronos, timesfm, moment, TTMs, visionts, aimts, timemoe, flame}.
By pre-training on large-scale time series datasets, these models exhibit strong adaptability to new tasks, enabling both efficient fine-tuning and zero-shot transfer across domains.
For instance, Timer~\citep{timer} employs a decoder-only architecture with autoregressive pre-training to capture temporal dependencies, while MOIRAI~\citep{moirai} introduces multi-scale patch projections to model diverse patterns and an any-variate attention mechanism that allows flexible handling of time series with arbitrary dimensionality.
ROSE~\citep{rose} combines frequency decomposition with registers to jointly learn both domain-invariant and domain-specific representations, facilitating knowledge transfer to downstream tasks. Sundial~\citep{sundial} proposes a TimeFlow Loss that predicts the distribution of the next patch, enabling Transformer training supporting probabilistic forecasting.

Existing TSFMs are all pre-trained solely on unimodal time series data, which provides some generalization ability but cannot leverage complementary modalities to model more complex temporal dynamics. In contrast, HORAI effectively leverages multiple modalities through a frequency-enhanced cross-modality encoder and introduces a Time-Frequency Decoder to further strengthen cross-modality and cross-domain generalization during pre-training.

\section{Methodology}

\subsection{Large-Scale Multimodal Time Series Dataset}

\begin{figure}[hbp]
    \centering 
\includegraphics[width=1\linewidth]{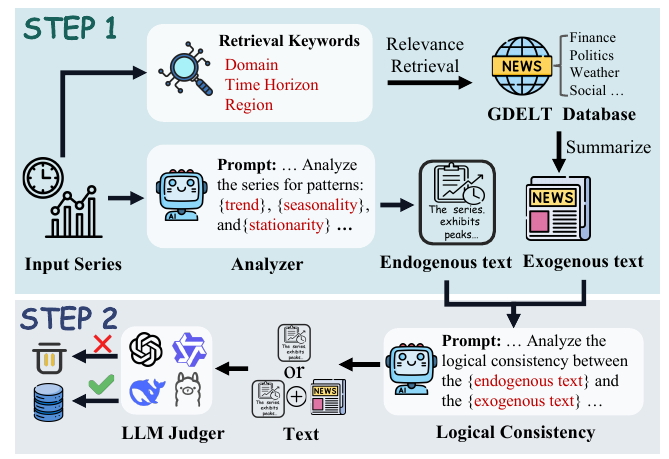}
\vspace{-1em}
\caption{
The automated data construction pipeline for multimodal text. It comprises two main stages: 1) Contextual Synthesis, involving endogenous pattern analysis and exogenous news retrieval; and 2) Quality Alignment, which ensures consistency between the synthesized texts and filters low-quality data via an LLM judger.
}
\label{figs:dataset generation}
\end{figure}

Large-scale datasets are the cornerstone of pre-training foundation models, enabling them to acquire transferable knowledge and improve generalization across diverse downstream scenarios. However, existing large-scale time series corpora are mostly confined to unimodal time series, which limits the potential of multimodal learning.
To address this problem, we conduct MM-TS, the first large-scale multimodal time series dataset for pre-training. As shown in Figure \ref{figs:dataset}, MM-TS integrates three modalities: time series, text, and image, covering six diverse domains, including Energy, Healthcare, Web, Nature, Transport, and Economics. In total, MM-TS contains over 1 billion time points.

For the time series modality, MM-TS spans multiple temporal granularities, including seconds, minutes, hours, and months, and captures diverse characteristics such as periodicity, trends, and non-stationarity (see Appendix \ref{pretrain datasets} for details). For the visual modality, we construct line-plot images directly from time series, offering an intuitive view of temporal fluctuations and structural patterns.

For the textual modality, to address the scarcity of high-quality, semantically aligned time series and textual pairs, we devise an automated data construction pipeline that uniquely integrates endogenous text with exogenous news. Endogenous text captures the semantic features of time series patterns, and exogenous news provides supplementary background context. Figure \ref{figs:dataset generation} shows that the pipeline includes two main stages: Contextual Synthesis and Quality Alignment. 
In the Contextual Synthesis stage, we adopt a dual-source generation strategy. Given a time series sample, an LLM analyzer GPT-4o first analyzes intrinsic patterns, such as trends and seasonalities, and converts them into structured semantic descriptions as Endogenous text. In parallel, we retrieve and summarize semantically related news events based on keywords from the GDELT Database to serve as Exogenous text.
Subsequently, the Quality Alignment stage ensures logical coherence and generation fidelity via two substeps: 1) Logical Consistency Filtering, where GPT-4o discards exogenous news conflicting with endogenous patterns to prevent spurious correlations; and 2) Ensemble Quality Evaluation, where the unified text is scored by multiple LLM judges on plausibility and relevance, retaining only high-confidence text. Further details on the text generation are in the Appendix \ref{MM-TS of Text Modality}.

\subsection{HORAI}
\begin{figure*}[t]
    \centering  
\includegraphics[width=1\linewidth]{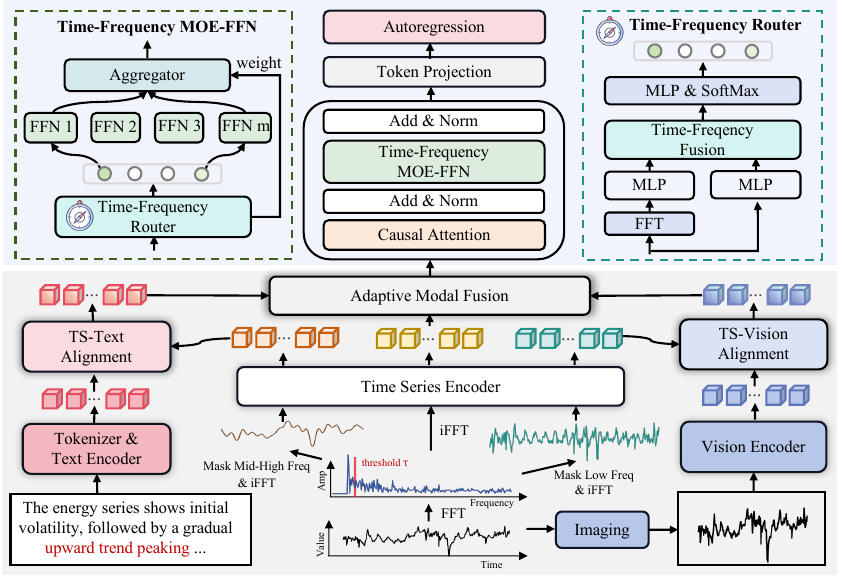}
\vspace{-1em}
\caption{The framework of the proposed HORAI consists of a Frequency-Enhanced Cross-Modal Encoder (gray region) and a Time-Frequency Decoder (blue region).}
\label{figs:Framework}
\vspace{-1em}
\end{figure*}

To better leverage cross-modal and cross-domain features for time series understanding, we propose HORAI, a frequency-enhanced multimodal foundation model for time series analysis. It consists of two core components: the Frequency-guided Cross-Modality Encoder and the Time-Frequency Decoder. As illustrated in Figure~\ref{figs:Framework}, in the Cross-Modality Encoder, the input time series is first decomposed into low-frequency and mid-to-high-frequency components, which are aligned with textual and visual features, respectively. Then, an adaptive modality fusion module combines these aligned representations to produce unified multimodal representations.
In the Time-Frequency Decoder, the multimodal representations are first passed into a Time-Frequency MoE-FFN, which is designed to capture diverse patterns across multiple domains. To guide the routing of tokens to appropriate experts, both temporal-domain and frequency-domain features are incorporated. The inclusion of frequency information provides additional cues that help distinguish similar patterns and group them coherently, enhancing the cross-modality and cross-domain generalization. Finally, the representations are projected through a token projection layer for autoregressive pre-training.

\subsubsection{Frequency-Enhanced Cross-Modal Encoder}

\paragraph{Multimodal Embedding.} 
For notational simplicity, we describe the method using a univariate time series, which can be easily extended to the multivariate case by treating each channel independently. Given an input time series $\mathbf{X}_{\mathrm{ts}} \in \mathbb{R}^{T}$, where $T$ denotes the sequence length, we first apply instance normalization \citep{revin} to mitigate distribution shift, resulting in $\mathbf{X}_\mathrm{norm} \in \mathbb{R}^{T}$. 

Since different frequency components capture different aspects of temporal dynamics, with low-frequency components reflecting global trends and mid-to-high-frequency components capturing local variations, we transform the normalized sequence into the frequency domain by the Fast Fourier Transform (FFT), obtaining $\mathbf{X}_\mathrm{freq} \in \mathbb{R}^{L/2+1}$. To separate different frequency bands, we set a ratio parameter $\alpha$ to define a cutoff threshold $\tau = \alpha \cdot (L/2+1)$. Based on this threshold, we construct two binary masks: $\mathbf{M}_\mathrm{low} \in \{0,1\}^{L/2+1}$ for low-frequency components and $\mathbf{M}_\mathrm{mh} \in \{0,1\}^{L/2+1}$ for mid-to-high-frequency components. 
Applying these masks to $\mathbf{X}_\mathrm{freq}$ by element-wise multiplication, which are then transformed back into the time domain using the inverse FFT (iFFT). This process produces the low-frequency sequence $\mathbf{X}_\mathrm{low} \in \mathbb{R}^{L}$ and the mid-to-high-frequency sequence $\mathbf{X}_\mathrm{mh} \in \mathbb{R}^{L}$.
\begin{equation}
    \begin{aligned}
        \mathbf{X}_{\mathrm{low}} = \mathrm{iFFT}(\mathbf{X}_{\mathrm{freq}} \odot \mathbf{M}_{\mathrm{mh}}), \\
        \mathbf{X}_{\mathrm{mh}} = \mathrm{iFFT}(\mathbf{X}_{\mathrm{freq}} \odot \mathbf{M}_{\mathrm{low}}). 
    \end{aligned}
\end{equation}
Subsequently, we employ a patching strategy to divide $\mathbf{X}_{\mathrm{low}}$, $\mathbf{X}_{\mathrm{mh}}$, and $\mathbf{X}_{\mathrm{norm}}$ into $N_{ts}$ patches with patch size $S$. These patches are projected and fed into the time-series encoder \citep{patchtst}, producing corresponding time series representations: $\mathbf{E}_\mathrm{low}$, $\mathbf{E}_\mathrm{mh}$, and $\mathbf{E}_\mathrm{ts} \in \mathbb{R}^{N_{ts} \times D_{ts}}$.

For the textual input $\mathbf{X}_{\mathrm{text}} \in \mathbb{R}^{L_{\mathrm{text}}}$, we employ a text tokenizer followed by a pre-trained text encoder Qwen-0.5B \cite{qwen} to extract semantic features, yielding $\mathbf{E}_{\mathrm{text}} \in \mathbb{R}^{L_{\mathrm{text}} \times D_{\mathrm{text}}}$. For the visual input $\mathbf{X}_{\mathrm{img}}$ 
, we apply a patching strategy and a pre-trained vision encoder ViT-Base \cite{ViT} to obtain image representations $\mathbf{E}_{\mathrm{img}} \in \mathbb{R}^{N_{\mathrm{img}} \times D_{\mathrm{img}}}$.

\paragraph{Frequency-enhanced Cross-Modality Alignment.} 

Time series often exhibit rich frequency-dependent patterns, where low-frequency components capture global trends and mid-to-high-frequency components reflect local variations. Meanwhile, different modalities contribute differently to these patterns: textual information tends to describe global trends, aligning with low-frequency time series components, whereas visual information focuses more on short-term variation, corresponding to mid-to-high-frequency components~\citep{timevlm}. Motivated by this, we propose a frequency-enhanced cross-modal fusion that explicitly leverages the characteristic correspondence between modalities and frequency components.
Additionally, given the large number of tokens in text and image modalities, we further integrate a Flow-Attention-based alignment mechanism to efficiently model cross-modal interactions.

In the TS-Text Alignment module, the low-frequency time series embeddings and textual embeddings are first projected by MLPs into a shared representation space $D_{model}$. Cross-modal fusion is then performed efficiently using the Flow-Attention mechanism \cite{flowformer}. The core idea is to treat attention as a flow of information and leverage the flow conservation principle to optimize the transmission and aggregation of features across modalities. Specifically, the low-frequency time series embeddings $\mathbf{E}_{\mathrm{low}}$ are mapped to serve as the Query $\mathbf{Q}$, while the textual embeddings $\mathbf{E'}_{\mathrm{text}}$ are mapped to serve as the Key $\mathbf{K}$ and Value $\mathbf{V}$. The information flow between tokens is computed as:
\begin{equation}
    \begin{aligned}
        & \mathbf{I}_i = \phi(\mathbf{Q}_i) \sum_{j=1}^{N_{\mathrm{text}}} \phi(\mathbf{K}_j)^T, \\ 
        \mathbf{\mathbf{O}}_j = \phi(\mathbf{K}_j) & \sum_{i=1}^{N_{\mathrm{ts}}} \phi (\mathbf{Q}_i)^T, 
        \hat{\mathbf{O}} = \phi(\mathbf{K}) \sum_{i=1}^{N_{\mathrm{ts}}} \frac{\phi(\mathbf{Q}_i)^T}{\mathbf{I}_i},\\
         \mathbf{E'}_{\mathrm{text}} = &\frac{\phi(\mathbf{Q})}{\mathbf{I}}(\phi(\mathbf{K})^T(\mathrm{Softmax}(\hat{\mathbf{O}})\odot \mathbf{V})),
    \end{aligned}
\end{equation}

$\phi(\cdot)$ denotes the non-linear projection to the flow space, $\mathbf{I}_i$ and $\mathbf{O}_j$ represent the total outgoing and incoming flows for each token. The output $\mathbf{E'}_{\mathrm{text}}$ 
is a flow-attention enhanced textual embedding, which has been adaptively aligned with the low-frequency time-series features.

Similar to the low-frequency time-series and text fusion, the TS-Vision Alignment module also leverages the Flow-Attention mechanism to integrate mid-to-high-frequency time-series embeddings $\mathbf{E}_{\mathrm{mh}}$ with image embeddings $\mathbf{E}_{\mathrm{img}}$, yielding aligned image representations $\mathbf{E'}_{\mathrm{img}} \in \mathbb{R}^{N_{\mathrm{ts}} \times D_{model}}$ for subsequent multimodal fusion.

\paragraph{Adaptive Modal Fusion.} Considering that the contributions of image and text representations vary across different time series patterns, we adaptively fuse the aligned image and text embeddings. The aligned image embeddings $\mathbf{E'}_\mathrm{img}$ and text embeddings $\mathbf{E'}_\mathrm{text}$ are concatenated along the feature dimension and then passed through a linear projection followed by a sigmoid function $\sigma$ to perform gated weighting $\mathbf{G}$, producing the multimodal representation $\mathbf{E}_{\mathrm{mm}}$. This representation is subsequently added to the time series embeddings $\mathbf{E}_{\mathrm{ts}}$ to obtain the fused representation $\mathbf{E}_\mathrm{fused} \in \mathbb{R}^{N_{ts} \times D_{model}}$. 
\begin{equation}
    \begin{aligned}
        &\mathbf{G} = \sigma(W_g[\mathbf{E'}_{\mathrm{img}}, \mathbf{E'}_\mathrm{{text}}] + b_g), \\
    \end{aligned}
\end{equation}
\begin{equation}
    \begin{aligned}
        \mathbf{E}_{\mathrm{fused}} &= \mathbf{G} \odot \mathbf{E'}_{\mathrm{img}} + (1-\mathbf{G}) \odot \mathbf{E'}_{\mathrm{text}} + \mathbf{E}_{\mathrm{ts}}.
    \end{aligned}
\end{equation}

\subsubsection{Time-Frequency Decoder}
Large-scale time series data inevitably involves diverse domains, which gives rise to a wide variety of temporal patterns \citep{rose, moirai}. The incorporation of textual and visual modalities further amplifies the diversity. To address this challenge, we propose a Time-Frequency Decoder designed to capture and adapt to different patterns, enhancing the generalization ability of pre-trained models. As illustrated in Figure~\ref{figs:Framework}, the Time-Frequency Decoder consists of key components including Causal Attention, Normalization layers, and a Time-Frequency MoE-FFN.

\paragraph{Time-Frequency MoE-FFN.} 
Different expert networks can capture distinct patterns from large-scale data, so effectively routing multimodal features to the appropriate experts is crucial. However, relying only on temporal-domain features may lead to entangled representations across different patterns, which makes pattern discrimination less straightforward. By incorporating frequency-domain features, similar patterns can be represented more compactly, offering additional cues for more accurate expert routing. Motivated by this, we propose the Time-Frequency Router, which integrates both temporal and frequency information to enhance the routing process.

Based on the fused multi-modal representation $\mathbf{E}_{\mathrm{fused}}$, we obtain representation $\mathbf{H}$ through causal attention followed by normalization. In the router, each token of $\mathbf{H}$ is projected in parallel across both temporal and frequency domains: (i) an MLP produces temporal representations $\mathbf{H}_{\mathrm{temp}}$, while (ii) an FFT followed by an MLP yields frequency representations $\mathbf{H}_{\mathrm{freq}}$. These dual-domain signals are adaptively integrated via a learnable gating function $G_{\text{router}}$, resulting in router representation $\mathbf{H}_r \in \mathbb{R}^{N_{ts} \times D_{model}}$:
\begin{equation}
    \begin{aligned}
\mathbf{H}_{\mathrm{r}}^i = \mathbf{G}_{\text{router}} \odot \mathrm{MLP}(\mathbf{H}^{i}) + (1-\mathbf{G}_{\text{router}}) \odot \mathrm{MLP}(\mathrm{FFT}(\mathbf{H}^{i})).
    \end{aligned}
\end{equation}
Given $\mathbf{H}_{r}$, the router applies an MLP-based routing function to compute routing weights $\mathbf{W} \in \mathbb{R}^{M}$, which determine expert assignment. Following a Top-K strategy, the router selects the $K$ experts with the highest weights, denoting the set of their indexes as $\mathcal{K}$. Then their outputs are aggregated through weight-normalized fusion, producing the representation $\mathbf{H}_{\mathrm{moe}} \in \mathbb{R}^{N_{ts} \times D_{model}}$:
\begin{equation}
    \begin{aligned}
\mathbf{H}_{\mathrm{moe}}^{i} = \sum_{j \in \mathcal{K}} \frac{\exp(\mathbf{W}_j)}{\sum_{m \in \mathcal{K}} \exp(\mathbf{W}_m)} \, \mathrm{FFN}_j(\mathbf{H}^{i}), \quad i = 1, \cdots, N_{ts}.
    \end{aligned}
\end{equation}

\paragraph{Autoregressive Training.}
Given the strong performance of the autoregressive paradigm in both NLP \cite{qwen, gpt3} and time series domains~\citep{sundial, timer}, we adopt a GPT-style training objective to predict the next token. This autoregressive formulation not only supports variable input and output lengths flexibly during inference but also excels at iterative, multi-step generation. Specifically, each input token $\mathbf{X}_{i} \in \mathbb{R}^{S}$ is processed through the encoder, decoder, and token projection layer to generate the prediction of the subsequent token $\hat{\mathbf{X}}_{i+1} \in \mathbb{R}^{S}$. The overall optimization objective is defined as:
\begin{equation}
    \begin{aligned}
        \mathcal{L}_{train} = \frac{1}{N_{ts}S} \sum ||\hat{\mathbf{X}}_i - \mathbf{X}_i||^2, \quad i =1, \dots, N_{ts}.
    \end{aligned}
\end{equation}

\section{Experiments}
\subsection{Experimental Setup}
\paragraph{Datasets.}    We perform pre-training of HORAI on our proposed MM-TS dataset and \textit{ensure no overlap between the pre-training MM-TS dataset and the downstream evaluation datasets}. To assess HORAI’s capability for time series analysis, we use the widely used evaluation datasets \citep{timemmd} for forecasting and anomaly detection tasks.
Specific dataset information is in Appendix \ref{datasets}.

\paragraph{Baselines.}   We select both time series foundation models and time-series-specific models as baselines. \textit{For the forecasting task}, we select five SOTA foundation models: 
ChatTime \citep{chattime}, VisionTS \citep{visionts}, ROSE \citep{rose}, Timer \citep{timer}, MOIRAI \citep{moirai}, and four \textit{multimodal time-series-specific models}: GPT4MTS \citep{gpt4mts}, TATS \citep{tats}, GPT4TS \citep{gpt4ts}, TimeVLM \citep{timevlm}. \textit{For the anomaly detection task}, we select three unimodal foundation models: DADA \citep{dada}, Timer, UniTS \citep{units}, and nine time-series-specific models: GPT4TS, LLMMixer \citep{llmmixer}, TimesNet \citep{Timesnet}, DCdetector \citep{DCdetector}, Anomlay Transformer(A.T.) \citep{A-T}, PatchTST \citep{patchtst}, HBOS \citep{HBOS}, IForest \citep{IForest}, and PCA \citep{pca}.

\begin{table*}[htbp]
  \centering
  \caption{Time series forecasting results under zero-shot and full-shot settings, reported as the average across four prediction horizons. The best results are highlighted in \textbf{bold}, and the second-best results are \underline{underlined}. Full results are presented in the Table \ref{tab:forecasting_full}.}
  \renewcommand{\arraystretch}{1.5}
  \resizebox{\linewidth}{!}{
    \begin{tabular}{c|cccccccccccc|cccccccc}
    \hline
    \hline
\textbf{Type} & \multicolumn{12}{c|}{\textbf{\emoji{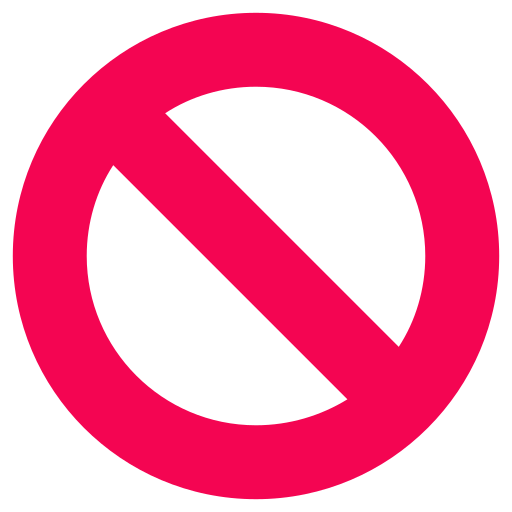} Time Series Foundation Models (Zero-Shot)}}                      & \multicolumn{8}{c}{\emoji{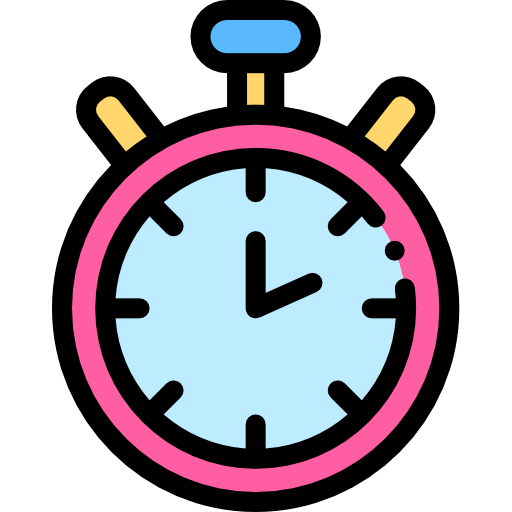} \textbf{Time-Series-Specific Models (Full-Shot)}} \\
    \hline
    \textbf{Models} & \multicolumn{2}{c}{\textbf{HORAI}} & \multicolumn{2}{c}{\textbf{ChatTime}} & \multicolumn{2}{c}{\textbf{VisionTS}} & \multicolumn{2}{c}{\textbf{ROSE}} & \multicolumn{2}{c}{\textbf{Timer}} & \multicolumn{2}{c|}{\textbf{MOIRAI}} & \multicolumn{2}{c}{\textbf{GPT4MTS}} & \multicolumn{2}{c}{\textbf{TATS}} & \multicolumn{2}{c}{\textbf{GPT4TS}} & \multicolumn{2}{c}{\textbf{TimeVLM}} \\
    \hline
    \textbf{Metric} & MSE   & MAE   & MSE   & MAE   & MSE   & MAE   & MSE   & MAE   & MSE   & MAE   & MSE   & MAE   & MSE   & MAE   & MSE   & MAE   & MSE   & MAE   & MSE   & MAE \\
    \hline
     \textbf{Agriculture} & \textbf{0.201 } & 0.309  & 0.369  & 0.410  & 0.290  & 0.336  & 0.345  & 0.372  & 0.289  & 0.339  & 0.272  & 0.403  & 0.225  & \underline{0.298 } & \underline{0.215 } & 0.301  & 0.220  & \textbf{0.294 } & 0.237  & 0.302  \\
    \textbf{Climate} & \textbf{0.857 } & \textbf{0.734 } & 1.860  & 1.106  & 1.307  & 0.930  & 1.475  & 0.987  & \underline{0.888 } & \underline{0.764 } & 1.921  & 1.095  & 1.182  & 0.889  & 1.180  & 0.887  & 1.184  & 0.891  & 1.195  & 0.899  \\
    \textbf{Energy} & \textbf{0.208 } & \textbf{0.325 } & \underline{0.247 } & \underline{0.352 } & 0.304  & 0.420  & 0.386  & 0.479  & 0.274  & 0.359  & 0.324  & 0.417  & 0.262  & 0.380  & 0.255  & 0.368  & 0.260  & 0.376  & 0.260  & 0.374  \\
    \textbf{Environment} & \textbf{0.309 } & \textbf{0.393 } & 0.359  & 0.456  & 0.354  & 0.436  & 0.392  & 0.456  & 0.351  & 0.428  & 0.351  & 0.403  & 0.323  & 0.400  & 0.319  & 0.396  & 0.322  & \underline{0.393 } & \underline{0.319 } & 0.397  \\
    \textbf{Social Good} & \textbf{0.819 } & 0.497  & 1.069  & 0.503  & 1.126  & 0.618  & 1.141  & 0.581  & 0.974  & 0.489  & 1.430  & 0.651  & 0.920  & 0.451  & 0.918  & \textbf{0.428 } & 0.917  & 0.476  & \underline{0.868 } & \underline{0.444 } \\
    \textbf{Traffic} & \textbf{0.165 } & \underline{0.244 } & 0.596  & 0.610  & 0.281  & 0.407  & 0.341  & 0.451  & 0.188  & 0.290  & 0.406  & 0.468  & 0.203  & 0.261  & \underline{0.179 } & \textbf{0.238 } & 0.206  & 0.266  & 0.216  & 0.319  \\
    \textbf{EWJ} & \textbf{0.589 } & \textbf{0.543 } & 0.887  & 0.641  & 0.645  & 0.584  & 0.706  & 0.605  & 0.696  & 0.595  & 0.937  & 0.688  & 0.626  & 0.549  & 0.612  & 0.546  & \underline{0.607 } & \underline{0.543 } & 0.609  & 0.544  \\
    \textbf{KR} & \textbf{0.551 } & \textbf{0.448 } & 0.565  & 0.455  & 0.671  & 0.522  & 0.555  & 0.480  & 0.549  & 0.463  & 0.992  & 0.629  & \underline{0.555 } & \underline{0.450 } & 0.578  & 0.449  & 0.578  & 0.448  & 0.584  & 0.454  \\
    \textbf{MDT} & \textbf{0.376 } & \textbf{0.434 } & 0.496  & 0.479  & 0.433  & 0.485  & 0.461  & 0.493  & 0.389  & 0.448  & 0.606  & 0.569  & \underline{0.385 } & 0.442  & 0.389  & \underline{0.436 } & 0.391  & 0.438  & 0.392  & 0.437  \\
    \rowc
    \multicolumn{1}{c|}{\textbf{1\textsuperscript{st} Count}} & \textbf{\textbf{9}} & \textbf{\textbf{6}} & 0 & 0 & 0 & 0 & 0 & 0 & 0 & 0 & 0 & 0 & 0 & 0 & 0 & \underline{2} & 0 & 1 & 0 & 0 \\
    \hline
    \hline
    \end{tabular}%
  }
  \label{tab:forecasting results}%
  \vspace{-1em}
\end{table*}%

\paragraph{Settings.} 
During pre-training, HORAI is optimized using the Adam optimizer with an initial learning rate of 0.0005 and trained for 20 epochs, employing an early stopping strategy with a patience of 5 epochs. 
For the forecasting task, all methods predict future values at four horizons to ensure a fair comparison. Additionally, \textit{none of the models employ the drop-last strategy} \citep{tfb}. For the anomaly detection task, evaluation is conducted using three score-based metrics: AUC-ROC, VUS-ROC, and VUS-PR \citep{vus_roc}, which are threshold-independent. Notably, \textbf{time series foundation models perform zero-shot inference directly, whereas time-series-specific models are trained in a full-shot setting for comparison}.

\subsection{Time Series Forecasting}

As shown in Table~\ref{tab:forecasting results}, HORAI achieves state-of-the-art forecasting performance compared to both unimodal foundation models and multimodal time-series-specific models, achieving top performance on 15 out of 18 cases. Specifically, relative to unimodal foundation models, HORAI 
outperforms ROSE with reductions of 29.6\% in MSE. These results indicate that HORAI effectively leverages multimodal information to enhance time series understanding and improve predictive accuracy. Compared to multimodal time-series-specific models trained in a full-shot manner, HORAI achieves superior performance even in the zero-shot setting: exceeding GPT4MTS by 11.4\% in MSE, and surpassing TimeVLM by 12.0\% in MSE. This demonstrates that pre-training on the large-scale multimodal time series dataset equips HORAI with strong generalization ability.

\subsection{Time Series Anomaly Detection}

As illustrated in Table~\ref{tab:anomalydetection results}, HORAI achieves state-of-the-art anomaly detection performance compared to both unimodal foundation models and time-series-specific models, attaining top results on 13 out of 15 cases. Compared to DADA, a general time series anomaly detector, HORAI outperforms it by 13.4\%, 19.5\%, and 19.2\% in AUC-ROC, VUS-ROC, and VUS-PR, respectively, under the zero-shot setting. This highlights that integrating multimodal data, such as text and images, enables the model to identify anomalous patterns better. Against time-series-specific models, HORAI outperforms GPT4TS by 12.2\%, 20.2\%, and 22.6\% in AUC-ROC, VUS-ROC, and VUS-PR. These results demonstrate that pre-training on large-scale, multi-domain data equips HORAI with general detection capability, effectively distinguishing between diverse normal and anomalous patterns.

\subsection{Ablation Study}

\begin{table*}[htbp]
  \centering
  \caption{Time series anomaly detection results under zero-shot and full-shot settings. The best results are in \textbf{bold}, and the second-best results are \underline{underlined}. More metric results are in Table \ref{tab: anomaly detection results with multiple metrics}.}
   \renewcommand{\arraystretch}{1}
  \resizebox{\linewidth}{!}{
    \begin{tabular}{c|c|cccc|ccccccccc}
    \hline
    \hline
    \textbf{Type} & \multicolumn{5}{c|}{\emoji{Figures/prohibit.png} \textbf{Time Series Foundation Models (Zero-Shot)}} & \multicolumn{9}{c}{\emoji{Figures/deadline.png} \textbf{Time-Series-Specific Models (Full-Shot)}} \\
    \midrule
    \textbf{Datasets} & \textbf{Metric} & \textbf{HORAI} & \textbf{DADA} & \textbf{Timer} & \textbf{UniTS} & \textbf{GPT4TS} & \textbf{LLMMixer} & \textbf{TimesNet} & \textbf{DCdetector} & \textbf{A.T.} & \textbf{PatchTST} & \textbf{HBOS} & \textbf{IForest} & \textbf{PCA} \\
    \midrule
    \multirow{3}[6]{*}{\textbf{EWJ}} & AUC\_ROC & \textbf{86.95} & 79.11 & 76.15 & 79.87 & 75.58 & 57.69  & \underline{82.39} & 53.40 & 43.81 & 78.53 & 71.82 & 69.20 & 54.35 \\
\cmidrule{2-15}          & VUS\_ROC & \textbf{82.78} & 71.79 & 67.72 & 73.91 & 67.95 & 52.79  & \underline{75.76} & 47.10 & 31.75 & 71.96 & 62.07 & 59.24 & 45.26 \\
\cmidrule{2-15}          & VUS\_PR & \textbf{48.27} & 43.36 & 33.17 & 39.32 & 35.63 & 15.13  & \underline{43.15} & 15.37 & 10.85 & 36.08 & 41.19 & 37.81 & 19.38 \\
    \midrule
    \multirow{3}[6]{*}{\textbf{MDT}} & AUC\_ROC & \textbf{91.22} & 79.04 & 75.65 & 73.19 & 74.79 & 60.30  & \underline{86.67} & 53.82 & 56.44 & 84.55 & 60.26 & 63.92 & 54.51 \\
\cmidrule{2-15}          & VUS\_ROC & \textbf{86.82} & 66.76 & 60.28 & 58.67 & 62.30 & 46.80  & \underline{83.40} & 45.02 & 44.53 & 77.69 & 55.30 & 54.02 & 44.09 \\
\cmidrule{2-15}          & VUS\_PR & \textbf{56.88} & 46.81 & 38.38 & 37.61 & 44.81 & 15.21  & \underline{52.13} & 15.72 & 15.93 & 41.67 & 44.77 & 35.32 & 22.93 \\
    \midrule
    \multirow{3}[6]{*}{\textbf{KR}} & AUC\_ROC & \textbf{96.38} & 79.53 & 66.72 & 80.95 & 78.30 & 65.77  & \underline{85.88} & 52.97 & 51.25 & 82.15 & 75.16 & 74.45 & 63.58 \\
\cmidrule{2-15}          & VUS\_ROC & \textbf{93.54} & 70.82 & 75.99 & 73.93 & 67.81 & 47.06  & \underline{79.00} & 43.04 & 41.97 & 74.65 & 58.77 & 60.70 & 47.51 \\
\cmidrule{2-15}          & VUS\_PR & \textbf{60.76} & 45.90 & 51.41 & 43.32 & 38.23 & 19.10  & \underline{51.60} & 8.49  & 7.94  & 36.18 & 54.17 & 43.31 & 24.19 \\
    \midrule
    \multirow{3}[6]{*}{\textbf{Energy}} & AUC\_ROC & \textbf{68.44} & 62.33 & 60.54 & 63.38 & 66.54 & 61.31  & \underline{68.36} & 48.75 & 38.68 & 66.70 & 60.80 & 60.32 & 61.14 \\
\cmidrule{2-15}          & VUS\_ROC & \textbf{62.42} & 54.37 & 46.03 & 51.15 & 53.10 & 53.04  & \underline{59.47} & 45.93 & 31.56 & 58.31 & 51.50 & 53.61 & 53.07 \\
\cmidrule{2-15}          & VUS\_PR & 35.24 & 34.18 & 29.46 & 31.04 & 31.68 & 30.35  & 38.61 & 22.57 & 19.69 & 34.41 & \underline{42.57} & \textbf{46.03} & 44.30 \\
    \midrule
    \multirow{3}[6]{*}{\textbf{Weather}} & AUC\_ROC & \textbf{81.49} & 66.37 & 80.86 & 81.22 & 74.47 & 79.60  & \underline{81.10} & 47.90 & 47.11 & 82.02 & 64.47 & 67.81 & 67.71 \\
\cmidrule{2-15}          & VUS\_ROC & \underline{80.40} & 61.03 & 73.22 & 75.08 & 70.03 & 71.71  & \textbf{81.91} & 45.56 & 43.32 & 79.97 & 54.16 & 56.45 & 57.38 \\
\cmidrule{2-15}          & VUS\_PR & \textbf{50.76} & 30.00 & 43.21 & 44.35 & 41.30 & 43.47  & 50.09 & 18.33 & 19.17 & \underline{50.13} & 46.58 & 49.66 & 47.13 \\
    \rowc
    \multicolumn{2}{c}{\textbf{1\textsuperscript{st} Count}} & \textbf{\textbf{13}} & 0 & 0 & 0 & 0 & 0 & 1 & 0 & 0 & 0 & 0 & 1 & 0  \\
    \hline
    \hline 
    \end{tabular}%
    }
  \label{tab:anomalydetection results}%
  \vspace{-1em}
\end{table*}%

\begin{figure}[htbp]
\vspace{-1em}
\includegraphics[width=1\linewidth]{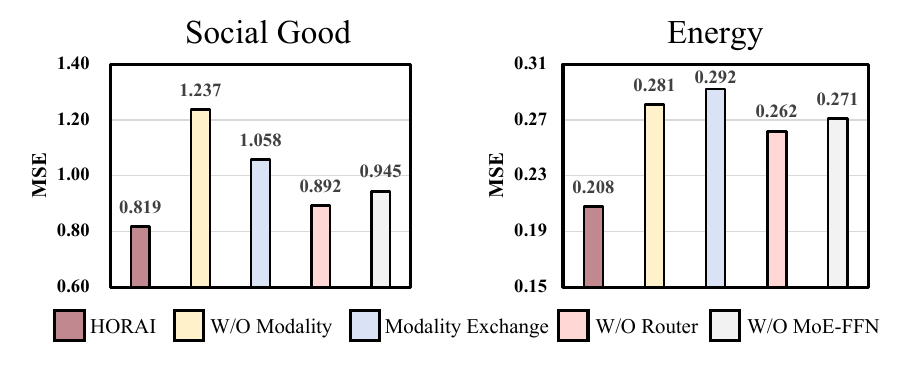}
\vspace{-2em}
\caption{Ablation study on the Social Good dataset and the Energy dataset.}
\vspace{-1em}
\label{figs:ablation}
\end{figure}

To evaluate the effectiveness of each component in HORAI, we conduct ablation experiments. Figure \ref{figs:ablation} illustrates the unique impact of each module. Removing the image and text modalities (W/O Modality) leads to a drop in performance, demonstrating that HORAI effectively leverages textual semantics and visual spatial information to enhance time series modeling. In the Modality Exchange variant, mid- and high-frequency time series features are aligned with texts, while low-frequency features are aligned with images. In contrast, HORAI aligns low-frequency features with text and mid- to high-frequency features with images, effectively exploiting the correspondence between modality-specific information and different frequency components of the time series, which improves modeling performance. 
This demonstrates that frequency-aware cross-modality alignment is crucial for capturing complementary patterns across modalities.
Replacing the Time-Frequency MoE-FFN with a standard FFN (W/O MoE-FFN) shows that the MoE-FFN allows each expert to capture distinct patterns, thereby enhancing the model’s generalization ability. Removing frequency information from the router (W/O Router) demonstrates that incorporating frequency information helps guide multimodal tokens to the most appropriate FFN experts, further improving performance.

\subsection{Model Analysis}
Due to the limitation of space, we provide the sensitivity analysis, analysis of specific modalities and alignment strategies, and experiments on replacing different text and vision encoders in Appendices \ref{sensitivity analysis}, \ref{specific modalities and alignment strategies}, and \ref{text encoder and vision encoder}, respectively.

\begin{figure}[htbp]
    \centering  
\includegraphics[width=1\linewidth]{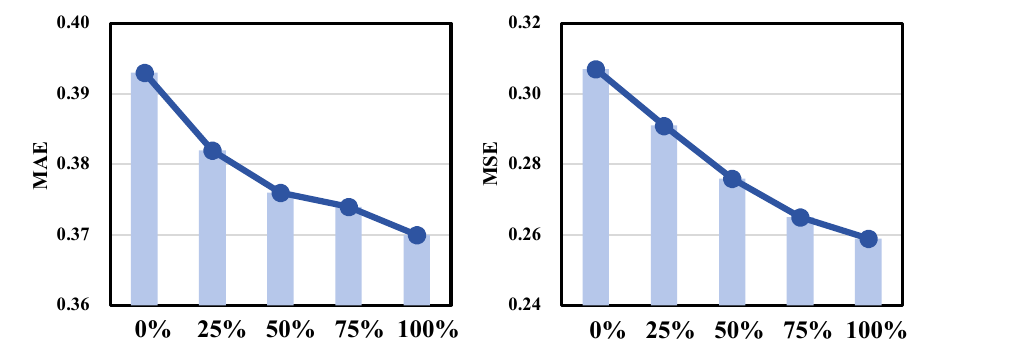}
\vspace{-1em}
\caption{Fine-tuning HORAI with different data percentages on the Environment dataset.}
\label{figs:modal_analysis}
\end{figure}

\paragraph{Fine-tune with downstream data.}
To examine how the amount of fine-tuning data affects downstream performance, we evaluate HORAI by progressively enlarging the training portion of the Environment dataset. As shown in Figure~\ref{figs:modal_analysis} (a), the forecasting accuracy steadily improves as more data is used, reaching its best with the full dataset. Specifically, the MAE decreases from 0.393 to 0.370, and the MSE decreases from 0.309 to 0.259. These results highlight HORAI’s strong adaptability to downstream data availability.

\begin{figure}[htbp]
    \centering
    \vspace{-1em}
    \includegraphics[width=1\linewidth]{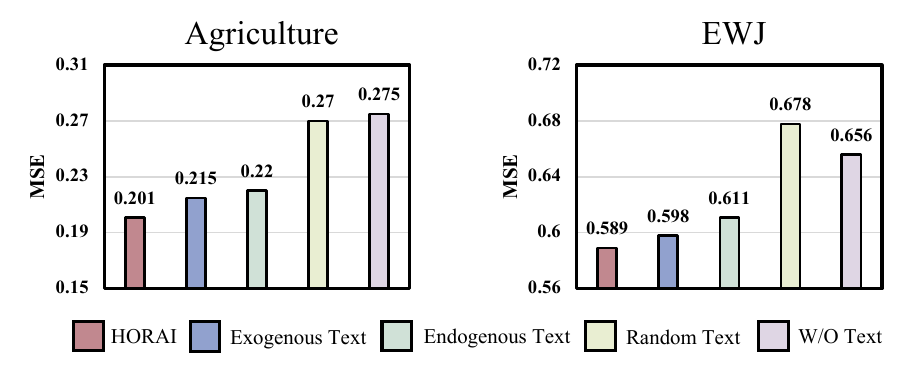}
    \vspace{-1em}
    \caption{Text replacement experiments on the Agriculture dataset and the EWJ dataset.}
    \vspace{-0.5em}
    \label{figs:text_replacement}
\end{figure}

\paragraph{Text Replacement.}
To examine whether HORAI genuinely leverages semantic information from two types of text, we conduct a series of text replacement experiments with four variants: randomly generated unrelated text (Random Text), exogenous text only (Exogenous Text), endogenous text only (Endogenous Text), and removing the entire text modality (W/O Text). As shown in Figure ~\ref{figs:text_replacement}, the full HORAI model consistently achieves the lowest MSE across both datasets, demonstrating the synergistic benefit of integrating endogenous descriptions with exogenous real-world news. Notably, utilizing either Exogenous Text or Endogenous Text independently yields better performance than W/O Text, confirming that both distinct sources provide valuable information. Introducing random text leads to a substantial performance drop, even worse than removing the text modality on the EWJ dataset, indicating that HORAI does not simply rely on the presence of text but actually understands and exploits its semantic content.

\section{Conclusion}
In this paper, we first propose a multimodal pretraining paradigm that leverages time series with endogenous images and text and exogenous news, providing a comprehensive multi-view perspective for time series analysis. To support this, we develop an automated data construction pipeline to curate MM-TS, the first large-scale multimodal time series dataset spanning six domains. Then we propose HORAI, a frequency-enhanced multimodal foundation model. It integrates two core components: the Frequency-guided Cross-Modality Encoder and the Time-Frequency Decoder, effectively fusing different multimodal features and enhancing model generalization. After pre-training on MM-TS, HORAI achieves SOTA performance in time series forecasting and anomaly detection tasks, which demonstrates strong task versatility and generalization ability.



\nocite{langley00}

\bibliography{example_paper}

@inproceedings{iTransformer,
  author       = {Yong Liu and
                  Tengge Hu and
                  Haoran Zhang and
                  Haixu Wu and
                  Shiyu Wang and
                  Lintao Ma and
                  Mingsheng Long},
  title        = {iTransformer: Inverted Transformers Are Effective for Time Series
                  Forecasting},
  booktitle    = {ICLR},
  year         = {2024}
}

@inproceedings{patchtst,
  author       = {Yuqi Nie and
                  Nam H. Nguyen and
                  Phanwadee Sinthong and
                  Jayant Kalagnanam},
  title        = {A Time Series is Worth 64 Words: Long-term Forecasting with Transformers},
  booktitle    = {ICLR},
  year         = {2023}
}

@inproceedings{timllm,
  author       = {Ming Jin and
                  Shiyu Wang and
                  Lintao Ma and
                  Zhixuan Chu and
                  James Y. Zhang and
                  Xiaoming Shi and
                  Pin{-}Yu Chen and
                  Yuxuan Liang and
                  Yuan{-}Fang Li and
                  Shirui Pan and
                  Qingsong Wen},
  title        = {Time-LLM: Time Series Forecasting by Reprogramming Large Language
                  Models},
  booktitle    = {ICLR},
  year         = {2024}
}

@inproceedings{gpt4ts,
  author       = {Tian Zhou and
                  Peisong Niu and
                  Xue Wang and
                  Liang Sun and
                  Rong Jin},
  title        = {One Fits All: Power General Time Series Analysis by Pretrained {LM}},
  booktitle    = {NeurIPS},
  year         = {2023}
}

@inproceedings{crossformer,
  author       = {Yunhao Zhang and
                  Junchi Yan},
  title        = {Crossformer: Transformer Utilizing Cross-Dimension Dependency for
                  Multivariate Time Series Forecasting},
  booktitle    = {ICLR},
  year         = {2023}
}

@inproceedings{dlinear,
  author       = {Ailing Zeng and
                  Muxi Chen and
                  Lei Zhang and
                  Qiang Xu},
  title        = {Are Transformers Effective for Time Series Forecasting?},
  booktitle    = {AAAI},
  pages        = {11121--11128},
  year         = {2023}
}

@inproceedings{pathformer,
  author       = {Peng Chen and
                  Yingying Zhang and
                  Yunyao Cheng and
                  Yang Shu and
                  Yihang Wang and
                  Qingsong Wen and
                  Bin Yang and
                  Chenjuan Guo},
  title        = {Pathformer: Multi-scale Transformers with Adaptive Pathways for Time
                  Series Forecasting},
  booktitle    = {ICLR},
  year         = {2024}
}

@inproceedings{fits,
  author       = {Zhijian Xu and
                  Ailing Zeng and
                  Qiang Xu},
  title        = {{FITS:} Modeling Time Series with 10k Parameters},
  booktitle    = {ICLR},
  year         = {2024}
}

@inproceedings{Timesnet,
  author       = {Haixu Wu and
                  Tengge Hu and
                  Yong Liu and
                  Hang Zhou and
                  Jianmin Wang and
                  Mingsheng Long},
  title        = {TimesNet: Temporal 2D-Variation Modeling for General Time Series Analysis},
  booktitle    = {ICLR},
  year         = {2023}
}

@article{deepar,
  author       = {Valentin Flunkert and
                  David Salinas and
                  Jan Gasthaus},
  title        = {DeepAR: Probabilistic Forecasting with Autoregressive Recurrent Networks},
  journal      = {CoRR},
  volume       = {abs/1704.04110},
  year         = {2017},
  eprinttype    = {arXiv}
}

@inproceedings{moderntcn,
  author       = {Donghao Luo and
                  Xue Wang},
  title        = {ModernTCN: {A} Modern Pure Convolution Structure for General Time
                  Series Analysis},
  booktitle    = {ICLR},
  year         = {2024}
}

@article{energy,
  author       = {Zhaoyang Zhu and
                  Weiqi Chen and
                  Rui Xia and
                  Tian Zhou and
                  Peisong Niu and
                  Bingqing Peng and
                  Wenwei Wang and
                  Hengbo Liu and
                  Ziqing Ma and
                  Xinyue Gu and
                  Jin Wang and
                  Qiming Chen and
                  Linxiao Yang and
                  Qingsong Wen and
                  Liang Sun},
  title        = {Energy forecasting with robust, flexible, and explainable machine
                  learning algorithms},
  journal      = {{AI} Mag.},
  volume       = {44},
  number       = {4},
  pages        = {377--393},
  year         = {2023}
}

@inproceedings{distillation,
  author       = {Mary Phuong and
                  Christoph Lampert},
  editor       = {Kamalika Chaudhuri and
                  Ruslan Salakhutdinov},
  title        = {Towards Understanding Knowledge Distillation},
  booktitle    = {ICML},
  volume       = {97},
  pages        = {5142--5151},
  year         = {2019}
}

@article{distillation1,
  author       = {Yoon Kim and
                  Alexander M. Rush},
  title        = {Sequence-Level Knowledge Distillation},
  journal      = {CoRR},
  volume       = {abs/1606.07947},
  year         = {2016},
  eprinttype    = {arXiv}
}

@inproceedings{gpt3,
  author       = {Tom B. Brown and
                  Benjamin Mann and
                  Nick Ryder and
                  Melanie Subbiah and
                  Jared Kaplan and
                  Prafulla Dhariwal and
                  Arvind Neelakantan and
                  Pranav Shyam and
                  Girish Sastry and
                  Amanda Askell and
                  Sandhini Agarwal and
                  Ariel Herbert{-}Voss and
                  Gretchen Krueger and
                  Tom Henighan and
                  Rewon Child and
                  Aditya Ramesh and
                  Daniel M. Ziegler and
                  Jeffrey Wu and
                  Clemens Winter and
                  Christopher Hesse and
                  Mark Chen and
                  Eric Sigler and
                  Mateusz Litwin and
                  Scott Gray and
                  Benjamin Chess and
                  Jack Clark and
                  Christopher Berner and
                  Sam McCandlish and
                  Alec Radford and
                  Ilya Sutskever and
                  Dario Amodei},
  title        = {Language Models are Few-Shot Learners},
  booktitle    = {NeurIPS},
  year         = {2020}
}

@article{tfb,
  author       = {Xiangfei Qiu and
                  Jilin Hu and
                  Lekui Zhou and
                  Xingjian Wu and
                  Junyang Du and
                  Buang Zhang and
                  Chenjuan Guo and
                  Aoying Zhou and
                  Christian S. Jensen and
                  Zhenli Sheng and
                  Bin Yang},
  title        = {{TFB:} Towards Comprehensive and Fair Benchmarking of Time Series
                  Forecasting Methods},
  journal      = {Proc. {VLDB} Endow.},
  volume       = {17},
  number       = {9},
  pages        = {2363--2377},
  year         = {2024}
}

@article{zhao2023multiple,
  author       = {Kai Zhao and
                  Chenjuan Guo and
                  Yunyao Cheng and
                  Peng Han and
                  Miao Zhang and
                  Bin Yang},
  title        = {Multiple Time Series Forecasting with Dynamic Graph Modeling},
  journal      = {Proc. {VLDB} Endow.},
  volume       = {17},
  number       = {4},
  pages        = {753--765},
  year         = {2023}
}

@inproceedings{ViT,
  author       = {Alexey Dosovitskiy and
                  Lucas Beyer and
                  Alexander Kolesnikov and
                  Dirk Weissenborn and
                  Xiaohua Zhai and
                  Thomas Unterthiner and
                  Mostafa Dehghani and
                  Matthias Minderer and
                  Georg Heigold and
                  Sylvain Gelly and
                  Jakob Uszkoreit and
                  Neil Houlsby},
  title        = {An Image is Worth 16x16 Words: Transformers for Image Recognition
                  at Scale},
  booktitle    = {ICLR},
  year         = {2021}
}

@inproceedings{moirai,
  author       = {Gerald Woo and
                  Chenghao Liu and
                  Akshat Kumar and
                  Caiming Xiong and
                  Silvio Savarese and
                  Doyen Sahoo},
  title        = {Unified Training of Universal Time Series Forecasting Transformers},
  booktitle    = {ICML},
  year         = {2024}
}

@inproceedings{timemoe,
  author       = {Xiaoming Shi and
                  Shiyu Wang and
                  Yuqi Nie and
                  Dianqi Li and
                  Zhou Ye and
                  Qingsong Wen and
                  Ming Jin},
  title        = {Time-MoE: Billion-Scale Time Series Foundation Models with Mixture
                  of Experts},
  booktitle      = {ICLR},
  year         = {2025}
}

@inproceedings{timer,
  author       = {Yong Liu and
                  Haoran Zhang and
                  Chenyu Li and
                  Xiangdong Huang and
                  Jianmin Wang and
                  Mingsheng Long},
  title        = {Timer: Generative Pre-trained Transformers Are Large Time Series Models},
  booktitle    = {ICML},
  year         = {2024}
}

@inproceedings{units,
  author       = {Shanghua Gao and
                  Teddy Koker and
                  Owen Queen and
                  Tom Hartvigsen and
                  Theodoros Tsiligkaridis and
                  Marinka Zitnik},
  title        = {UniTS: {A} Unified Multi-Task Time Series Model},
  booktitle    = {NeurIPS},
  year         = {2024}
}

@inproceedings{timesfm,
  author       = {Abhimanyu Das and
                  Weihao Kong and
                  Rajat Sen and
                  Yichen Zhou},
  title        = {A decoder-only foundation model for time-series forecasting},
  booktitle    = {ICML},
  year         = {2024}
}

@article{chronos,
  author       = {Abdul Fatir Ansari and
                  Lorenzo Stella and
                  Ali Caner T{\"{u}}rkmen and
                  Xiyuan Zhang and
                  Pedro Mercado and
                  Huibin Shen and
                  Oleksandr Shchur and
                  Syama Sundar Rangapuram and
                  Sebastian Pineda{-}Arango and
                  Shubham Kapoor and
                  Jasper Zschiegner and
                  Danielle C. Maddix and
                  Michael W. Mahoney and
                  Kari Torkkola and
                  Andrew Gordon Wilson and
                  Michael Bohlke{-}Schneider and
                  Yuyang Wang},
  title        = {Chronos: Learning the Language of Time Series},
  journal      = {CoRR},
  volume       = {abs/2403.07815},
  year         = {2024}
}

@inproceedings{moment,
  author       = {Mononito Goswami and
                  Konrad Szafer and
                  Arjun Choudhry and
                  Yifu Cai and
                  Shuo Li and
                  Artur Dubrawski},
  title        = {{MOMENT:} {A} Family of Open Time-series Foundation Models},
  booktitle    = {ICML},
  year         = {2024}
}

@article{anomly_survey,
  author       = {Zahra Zamanzadeh Darban and
                  Geoffrey I. Webb and
                  Shirui Pan and
                  Charu Aggarwal and
                  Mahsa Salehi},
  title        = {Deep Learning for Time Series Anomaly Detection: {A} Survey},
  journal      = {{ACM} Comput. Surv.},
  volume       = {57},
  number       = {1},
  pages        = {15:1--15:42},
  year         = {2025}
}

@inproceedings{A-T,
  author       = {Jiehui Xu and
                  Haixu Wu and
                  Jianmin Wang and
                  Mingsheng Long},
  title        = {Anomaly Transformer: Time Series Anomaly Detection with Association
                  Discrepancy},
  booktitle    = {ICLR},
  year         = {2022}
}

@inproceedings{DCdetector,
  author       = {Yiyuan Yang and
                  Chaoli Zhang and
                  Tian Zhou and
                  Qingsong Wen and
                  Liang Sun},
  title        = {DCdetector: Dual Attention Contrastive Representation Learning for
                  Time Series Anomaly Detection},
  booktitle    = {SIGKDD},
  pages        = {3033--3045},
  year         = {2023}
}

@article{tsb,
  title={TSB-UAD: an end-to-end benchmark suite for univariate time-series anomaly detection},
  author={Paparrizos, John and Kang, Yuhao and Boniol, Paul and Tsay, Ruey S and Palpanas, Themis and Franklin, Michael J},
  journal={VLDB},
  volume={15},
  number={8},
  pages={1697--1711},
  year={2022}
}

@article{autocts,
  author       = {Xinle Wu and
                  Dalin Zhang and
                  Chenjuan Guo and
                  Chaoyang He and
                  Bin Yang and
                  Christian S. Jensen},
  title        = {AutoCTS: Automated Correlated Time Series Forecasting},
  journal      = {VLDB},
  volume       = {15},
  number       = {4},
  pages        = {971--983},
  year         = {2021}
}

@inproceedings{TTMs,
  author       = {Vijay Ekambaram and
                  Arindam Jati and
                  Pankaj Dayama and
                  Sumanta Mukherjee and
                  Nam Nguyen and
                  Wesley M. Gifford and
                  Chandra Reddy and
                  Jayant Kalagnanam},
  title        = {Tiny Time Mixers (TTMs): Fast Pre-trained Models for Enhanced Zero/Few-Shot
                  Forecasting of Multivariate Time Series},
  booktitle    = {NeurIPS},
  year         = {2024}
}

@article{monash,
  title={Monash time series forecasting archive},
  author={Godahewa, Rakshitha and Bergmeir, Christoph and Webb, Geoffrey I and Hyndman, Rob J and Montero-Manso, Pablo},
  journal={arXiv preprint arXiv:2105.06643},
  year={2021}
}

@article{ucr,
  title={The UCR time series archive},
  author={Dau, Hoang Anh and Bagnall, Anthony and Kamgar, Kaveh and Yeh, Chin-Chia Michael and Zhu, Yan and Gharghabi, Shaghayegh and Ratanamahatana, Chotirat Ann and Keogh, Eamonn},
  journal={IEEE/CAA Journal of Automatica Sinica},
  volume={6},
  number={6},
  pages={1293--1305},
  year={2019},
  publisher={IEEE}
}

@article{uea,
  title={The UEA multivariate time series classification archive, 2018},
  author={Bagnall, Anthony and Dau, Hoang Anh and Lines, Jason and Flynn, Michael and Large, James and Bostrom, Aaron and Southam, Paul and Keogh, Eamonn},
  journal={arXiv preprint arXiv:1811.00075},
  year={2018}
}

@article{tdbrain,
  title={Contrast everything: A hierarchical contrastive framework for medical time-series},
  author={Wang, Yihe and Han, Yu and Wang, Haishuai and Zhang, Xiang},
  journal={NeurIPS},
  volume={36},
  year={2024}
}

@article{prsa,
  title={Cautionary tales on air-quality improvement in Beijing},
  author={Zhang, Shuyi and Guo, Bin and Dong, Anlan and He, Jing and Xu, Ziping and Chen, Song Xi},
  journal={Proceedings of the Royal Society A: Mathematical, Physical and Engineering Sciences},
  volume={473},
  number={2205},
  pages={20170457},
  year={2017},
  publisher={The Royal Society Publishing}
}

@article{pems,
  title={Scinet: Time series modeling and forecasting with sample convolution and interaction},
  author={Liu, Minhao and Zeng, Ailing and Chen, Muxi and Xu, Zhijian and Lai, Qiuxia and Ma, Lingna and Xu, Qiang},
  journal={NeurIPS},
  volume={35},
  pages={5816--5828},
  year={2022}
}

@article{fred,
  title={FRED-MD: A monthly database for macroeconomic research},
  author={McCracken, Michael W and Ng, Serena},
  journal={Journal of Business \& Economic Statistics},
  volume={34},
  number={4},
  pages={574--589},
  year={2016}
}

@article{nn5,
  title={A review and comparison of strategies for multi-step ahead time series forecasting based on the NN5 forecasting competition},
  author={Taieb, Souhaib Ben and Bontempi, Gianluca and Atiya, Amir F and Sorjamaa, Antti},
  journal={Expert systems with applications},
  volume={39},
  number={8},
  pages={7067--7083},
  year={2012}
}

@article{gluonts,
  author       = {Alexander Alexandrov and
                  Konstantinos Benidis and
                  Michael Bohlke{-}Schneider and
                  Valentin Flunkert and
                  Jan Gasthaus and
                  Tim Januschowski and
                  Danielle C. Maddix and
                  Syama Sundar Rangapuram and
                  David Salinas and
                  Jasper Schulz and
                  Lorenzo Stella and
                  Ali Caner T{\"{u}}rkmen and
                  Yuyang Wang},
  title        = {GluonTS: Probabilistic and Neural Time Series Modeling in Python},
  journal      = {J. Mach. Learn. Res.},
  volume       = {21},
  pages        = {116:1--116:6},
  year         = {2020}
}

@article{SZ-Taxi,
  author       = {Jingyuan Wang and
                  Jiawei Jiang and
                  Wenjun Jiang and
                  Chengkai Han and
                  Wayne Xin Zhao},
  title        = {Towards Efficient and Comprehensive Urban Spatial-Temporal Prediction:
                  {A} Unified Library and Performance Benchmark},
  journal      = {CoRR},
  volume       = {abs/2304.14343},
  year         = {2023}
}

@article{vus_roc,
  title={Volume under the surface: a new accuracy evaluation measure for time-series anomaly detection},
  author={Paparrizos, John and Boniol, Paul and Palpanas, Themis and Tsay, Ruey S and Elmore, Aaron and Franklin, Michael J},
  journal={Proceedings of the VLDB Endowment},
  volume={15},
  number={11},
  pages={2774--2787},
  year={2022}
}

@article{timemmd,
  title={Time-mmd: Multi-domain multimodal dataset for time series analysis},
  author={Liu, Haoxin and Xu, Shangqing and Zhao, Zhiyuan and Kong, Lingkai and Prabhakar Kamarthi, Harshavardhan and Sasanur, Aditya and Sharma, Megha and Cui, Jiaming and Wen, Qingsong and Zhang, Chao and others},
  journal={NeurIPS},
  volume={37},
  pages={77888--77933},
  year={2024}
}

@inproceedings{internvl,
  title={Internvl: Scaling up vision foundation models and aligning for generic visual-linguistic tasks},
  author={Chen, Zhe and Wu, Jiannan and Wang, Wenhai and Su, Weijie and Chen, Guo and Xing, Sen and Zhong, Muyan and Zhang, Qinglong and Zhu, Xizhou and Lu, Lewei and others},
  booktitle={CVPR},
  pages={24185--24198},
  year={2024}
}

@inproceedings{next-gpt,
  author       = {Shengqiong Wu and
                  Hao Fei and
                  Leigang Qu and
                  Wei Ji and
                  Tat{-}Seng Chua},
  title        = {NExT-GPT: Any-to-Any Multimodal {LLM}},
  booktitle    = {ICML},
  year         = {2024}
}

@article{sundial,
  author       = {Yong Liu and
                  Guo Qin and
                  Zhiyuan Shi and
                  Zhi Chen and
                  Caiyin Yang and
                  Xiangdong Huang and
                  Jianmin Wang and
                  Mingsheng Long},
  title        = {Sundial: {A} Family of Highly Capable Time Series Foundation Models},
  journal      = {CoRR},
  volume       = {abs/2502.00816},
  year         = {2025},
  eprinttype    = {arXiv}
}

@inproceedings{rose,
  title={Towards a General Time Series Forecasting Model with Unified Representation and Adaptive Transfer},
  author={Wang, Yihang and Qiu, Yuying and Chen, Peng and Zhao, Kai and Shu, Yang and Rao, Zhongwen and Pan, Lujia and Yang, Bin and Guo, Chenjuan},
  booktitle={ICML},
  year = {2025}
}

@inproceedings{timevlm,
  title={Time-VLM: Exploring Multimodal Vision-Language Models for Augmented Time Series Forecasting},
  author={Zhong, Siru and Ruan, Weilin and Jin, Ming and Li, Huan and Wen, Qingsong and Liang, Yuxuan},
  booktitle={ICML},
  year={2025}
}

@inproceedings{revin,
  title={Reversible instance normalization for accurate time-series forecasting against distribution shift},
  author={Kim, Taesung and Kim, Jinhee and Tae, Yunwon and Park, Cheonbok and Choi, Jang-Ho and Choo, Jaegul},
  booktitle={ICLR},
  year={2021}
}

@article{intervention,
  title={Intervention-Aware Forecasting: Breaking Historical Limits from a System Perspective},
  author={Xu, Zhijian and Wang, Hao and Xu, Qiang},
  journal={arXiv preprint arXiv:2405.13522},
  year={2024}
}

@article{qwen,
  title={Qwen technical report},
  author={Bai, Jinze and Bai, Shuai and Chu, Yunfei and Cui, Zeyu and Dang, Kai and Deng, Xiaodong and Fan, Yang and Ge, Wenbin and Han, Yu and Huang, Fei and others},
  journal={arXiv preprint arXiv:2309.16609},
  year={2023}
}

@inproceedings{gpt4mts,
  title={Gpt4mts: Prompt-based large language model for multimodal time-series forecasting},
  author={Jia, Furong and Wang, Kevin and Zheng, Yixiang and Cao, Defu and Liu, Yan},
  booktitle={AAAI},
  volume={38},
  number={21},
  pages={23343--23351},
  year={2024}
}

@article{visionts,
  title={Visionts: Visual masked autoencoders are free-lunch zero-shot time series forecasters},
  author={Chen, Mouxiang and Shen, Lefei and Li, Zhuo and Wang, Xiaoyun Joy and Sun, Jianling and Liu, Chenghao},
  journal={arXiv preprint arXiv:2408.17253},
  year={2024}
}

@article{tats,
  title={Language in the flow of time: Time-series-paired texts weaved into a unified temporal narrative},
  author={Li, Zihao and Lin, Xiao and Liu, Zhining and Zou, Jiaru and Wu, Ziwei and Zheng, Lecheng and Fu, Dongqi and Zhu, Yada and Hamann, Hendrik and Tong, Hanghang and others},
  journal={arXiv preprint arXiv:2502.08942},
  year={2025}
}

@inproceedings{dada,
  title={Towards a General Time Series Anomaly Detector with Adaptive Bottlenecks and Dual Adversarial Decoders},
  author={Shentu, Qichao and Li, Beibu and Zhao, Kai and Shu, Yang and Rao, Zhongwen and Pan, Lujia and Yang, Bin and Guo, Chenjuan},
  booktitle={ICLR},
  year={2025}
}

@article{llmmixer,
  title={Llm-mixer: Multiscale mixing in llms for time series forecasting},
  author={Kowsher, Md and Sobuj, Md Shohanur Islam and Prottasha, Nusrat Jahan and Alanis, E Alejandro and Garibay, Ozlem Ozmen and Yousefi, Niloofar},
  journal={arXiv preprint arXiv:2410.11674},
  year={2024}
}

@article{cctime,
  title={CC-Time: Cross-Model and Cross-Modality Time Series Forecasting},
  author={Chen, Peng and Wang, Yihang and Shu, Yang and Cheng, Yunyao and Zhao, Kai and Rao, Zhongwen and Pan, Lujia and Yang, Bin and Guo, Chenjuan},
  journal={arXiv preprint arXiv:2508.12235},
  year={2025}
}

@article{pca,
  title={A novel anomaly detection scheme based on principal component classifier},
  author={Shyu, Mei-Ling and Chen, Shu-Ching and Sarinnapakorn, Kanoksri and Chang, LiWu},
  year={2003}
}

@inproceedings{IForest,
  title={Isolation forest},
  author={Liu, Fei Tony and Ting, Kai Ming and Zhou, Zhi-Hua},
  booktitle={ICDM},
  pages={413--422},
  year={2008}
}

@article{HBOS,
  title={Histogram-based outlier score (hbos): A fast unsupervised anomaly detection algorithm},
  author={Goldstein, Markus and Dengel, Andreas},
  journal={KI-2012: poster and demo track},
  volume={1},
  pages={59--63},
  year={2012}
}

@inproceedings{FNSPID,
  author       = {Zihan Dong and
                  Xinyu Fan and
                  Zhiyuan Peng},
  editor       = {Ricardo Baeza{-}Yates and
                  Francesco Bonchi},
  title        = {{FNSPID:} {A} Comprehensive Financial News Dataset in Time Series},
  booktitle    = {SIGKDD},
  pages        = {4918--4927},
  year         = {2024}
}

@inproceedings{chattime,
  author       = {Chengsen Wang and
                  Qi Qi and
                  Jingyu Wang and
                  Haifeng Sun and
                  Zirui Zhuang and
                  Jinming Wu and
                  Lei Zhang and
                  Jianxin Liao},
  title        = {ChatTime: {A} Unified Multimodal Time Series Foundation Model Bridging
                  Numerical and Textual Data},
  booktitle    = {AAAI},
  pages        = {12694--12702},
  year         = {2025}
}

@article{flowformer,
  title={Flowformer: Linearizing transformers with conservation flows},
  author={Wu, Haixu and Wu, Jialong and Xu, Jiehui and Wang, Jianmin and Long, Mingsheng},
  journal={arXiv preprint arXiv:2202.06258},
  year={2022}
}

@misc{flame,
title           = {FLAME: Flow Enhanced Legendre Memory Models for General Time Series Forecasting},
author          = {{Anonymous Authors}}, 
howpublished    = {Concurrent Submission to ICML},
year            = {2026},
note            = {Filename: flame\_ICML\_2026.pdf}
}

@article{dtaf,
  title={Towards Non-Stationary Time Series Forecasting with Temporal Stabilization and Frequency Differencing},
  author={Lu, Junkai and Chen, Peng and Guo, Chenjuan and Shu, Yang and Wang, Meng and Yang, Bin},
  journal={arXiv preprint arXiv:2511.08229},
  year={2025}
}

@article{aimts,
  title={AimTS: Augmented series and image contrastive learning for time series classification},
  author={Chen, Yuxuan and Huang, Shanshan and Cheng, Yunyao and Chen, Peng and Rao, Zhongwen and Shu, Yang and Yang, Bin and Pan, Lujia and Guo, Chenjuan},
  journal={arXiv preprint arXiv:2504.09993},
  year={2025}
}
\bibliographystyle{icml2026}

\newpage
\appendix
\onecolumn

\appendix
\section{Datasets}
\label{datasets}

\subsection{Pre-train Dataset MM-TS of Time Series Modality}
\label{pretrain datasets}
For time series modality, we assemble a large and diverse set of publicly available time series datasets covering domains such as energy, nature, transportation, web, health, and economics. The corpus contains around 1 billion time points, with a strict separation from all target evaluation datasets. The datasets vary widely in their sampling frequencies—from millisecond-level measurements to monthly observations—reflecting both the heterogeneity of real-world scenarios and the complexity of temporal dynamics.
\begin{table}[h]
  \centering
  \caption{List of pretraining datasets of time series modality.}
  \resizebox{0.75\linewidth}{!}{
    \begin{tabular}{c|c|c|c|c}
    \hline
    \hline
    \textbf{Domain} & \textbf{Dataset} & \textbf{Frequency} & \textbf{Time Pionts} & \textbf{Source} \\
    \midrule
    \multirow{8}[8]{*}{Energy} & Aus. Electricity Demand & Half Hourly & 1155264 & Monash~\citep{monash} \\
\cmidrule{2-5}          & Wind  & 4 Seconds & 7397147 & Monash~\citep{monash} \\
\cmidrule{2-5}          & Wind Farms & Minutely & 172178060 & Monash~\citep{monash} \\
\cmidrule{2-5}          & Solar Power & 4 Seconds & 7397222 & Monash~\citep{monash} \\
\cmidrule{2-5}          & London Smart Meters & Half Hourly & 166527216 & Monash~\citep{monash} \\
\cmidrule{2-5}          & BDG-2 Rat  &Hourly  & 4596080 & \citep{gluonts} \\
\cmidrule{2-5}          & BDG-2 Panther &Hourly & 893840 & \citep{gluonts} \\
\cmidrule{2-5}          & BDG-2 Fox &Hourly &2285288 & \citep{gluonts} \\
    \midrule
    \multirow{11}[22]{*}{Nature} & Phoneme & -     & 2160640 & UCR\cite{ucr} \\
\cmidrule{2-5}          & PRSA  & Hourly & 4628448 & ~\citep{prsa} \\
\cmidrule{2-5}          & Temperature Rain & Daily & 23252200 & Monash~\citep{monash} \\
\cmidrule{2-5}          & StarLightCurves & -     & 9457664 & UCR~\citep{ucr} \\
\cmidrule{2-5}          & Worms & 0.033 Seconds & 232200 & UCR~\citep{ucr} \\
\cmidrule{2-5}          & Saugeen River Flow & Daily & 23741 & Monash~\citep{monash} \\
\cmidrule{2-5}          & Sunspot & Daily & 73924 & Monash~\citep{monash} \\
\cmidrule{2-5}          & Weather & Daily & 43032000 & Monash~\citep{monash} \\
\cmidrule{2-5}          & KDD Cup 2018 & Daily & 2942364 & Monash\cite{monash} \\
\cmidrule{2-5}          & US Births & Daily & 7305  & Monash~\citep{monash} \\
    \midrule
    \multirow{4}[14]{*}{Healthcare} & MotorImagery & 0.001 Seconds & 72576000 & UEA~\citep{uea} \\
\cmidrule{2-5}          & AtrialFibrillation & 0.008 Seconds & 38400 & UEA~\citep{uea} \\
\cmidrule{2-5}          & PigArtPressure & -     & 624000 & UCR~\citep{ucr} \\
\cmidrule{2-5}          & TDbrain & 0.002 Seconds & 79232703 & ~\citep{tdbrain} \\
    \midrule
    \multirow{9}[12]{*}{Transport} & Pems03 & 5 Minute & 9382464 & ~\citep{pems} \\
\cmidrule{2-5}          & Pems08 & 5 Minute & 3035520 & ~\citep{pems} \\
\cmidrule{2-5}          & Pems-bay & 5 Minute & 16937700 & ~\citep{pems} \\
\cmidrule{2-5}          & Pedestrian\_Counts & Hourly & 3132346 & Monash~\citep{monash} \\
\cmidrule{2-5}          & SZ-Taxi & 15 Minute &464256 & \citep{SZ-Taxi}\\
\cmidrule{2-5}          & Taxi & Half Hourly &40584636 & \citep{gluonts} \\    
\cmidrule{2-5}          & Uber TLC & Hourly &510284 &  \citep{gluonts}\\
    \midrule
    Web   & Web Traffic & Daily & 116485589 & Monash~\citep{monash} \\
    \midrule
    \multirow{3}[6]{*}{Economic} & FRED\_MD & Monthly & 77896 & ~\citep{fred} \\
\cmidrule{2-5}          & Bitcoin & Daily & 75364 & Monash~\citep{monash} \\
\cmidrule{2-5}          & NN5   & Daily & 87801 & ~\citep{nn5} \\
    \hline
    \hline

    \end{tabular}}%
  \label{tab: pretraining datasets}%
\end{table}%

\subsection{Pre-train Dataset MM-TS of Text Modality}
\label{MM-TS of Text Modality}

To address the scarcity of high-quality, semantically aligned time series and text pairs, we design a two-stage automated data construction pipeline. The pipeline uniquely integrates endogenous text (time series patterns description) with exogenous real-world news events through a "Generate-then-Filter" paradigm.

\begin{figure}[hbp]
    \centering
    \vspace{-0.5em}
    \includegraphics[width=0.9\linewidth]{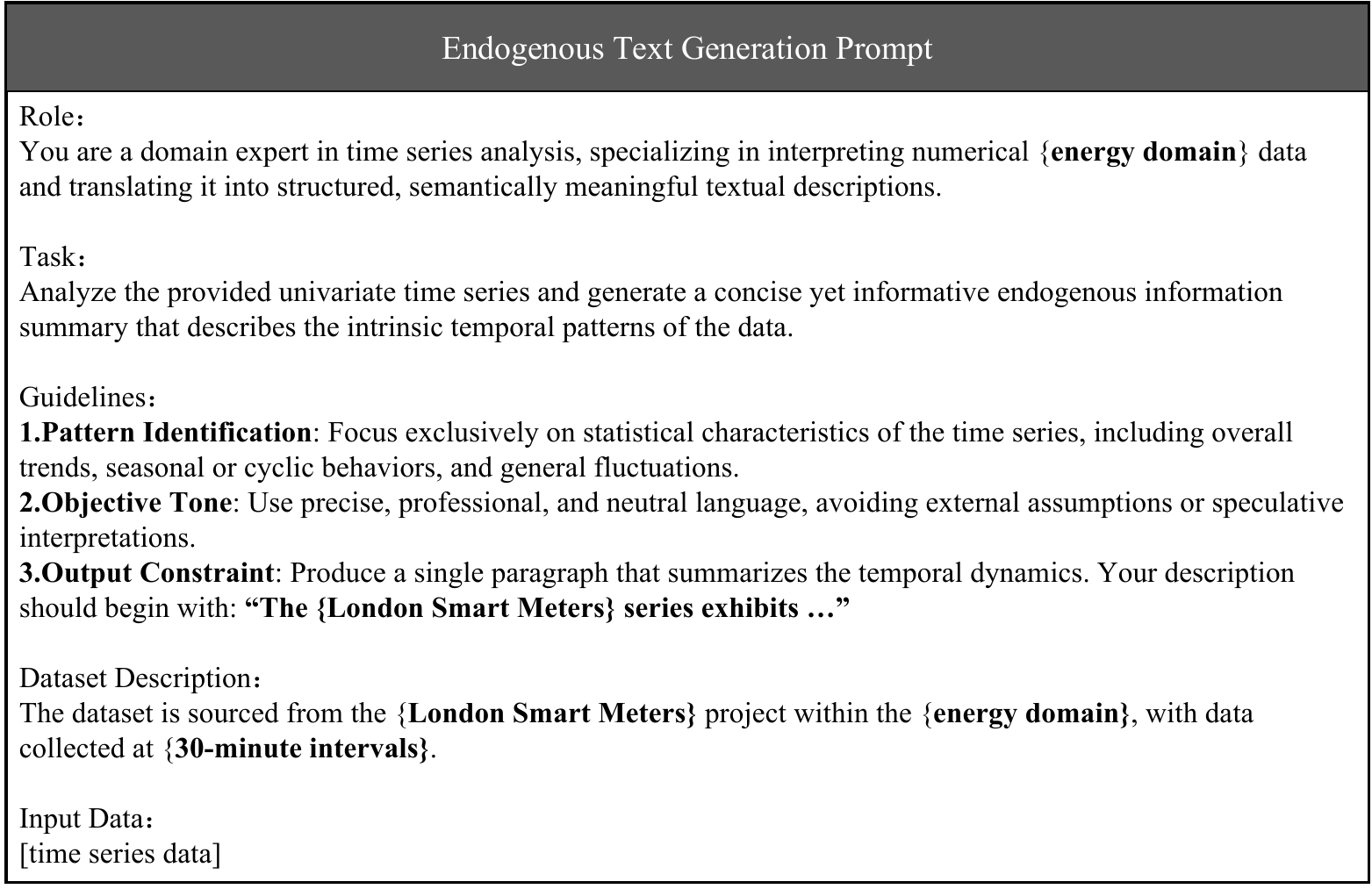}
    \caption{The prompt is designed for generating endogenous descriptions of the London Smart Meters dataset.}
    \label{fig:endogenous text}
\end{figure}

\begin{figure}[hbp]
    \centering
    \vspace{-1em}
    \includegraphics[width=0.9\linewidth]{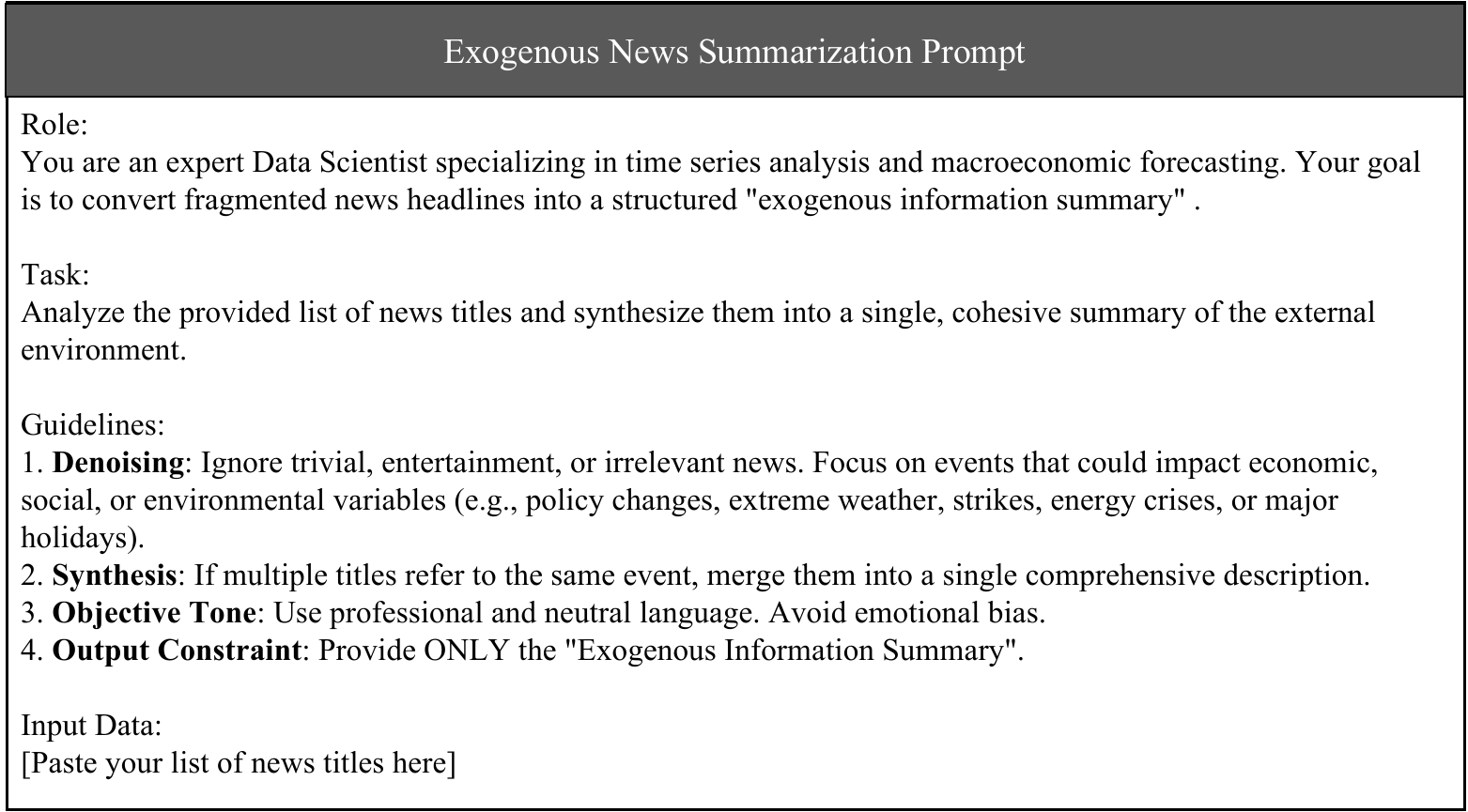}
    \caption{The prompt is designed for summarizing the relevant exogenous news.}
    \label{fig:exogenous text}
\end{figure}

Stage 1: Contextual Synthesis Stage.
We adopt a parallel generation strategy to synthesize the textual modality:

\begin{itemize}
    \item 
Endogenous Text Generation: An LLM Analyzer GPT-4o is employed to interpret the numerical time series. As shown in Figure \ref{fig:endogenous text}, guided by specific prompts targeting statistical properties (e.g., trend, seasonality, and stationarity), the analyzer converts raw temporal dynamics into structured Endogenous Text descriptions.
    \item
Exogenous News Retrieval: Simultaneously, we extract metadata constraints—specifically Domain, Time Horizon, and Region—to formulate search queries. We utilize a Relevance Retrieval to fetch the relevant news events from the GDELT Database. These raw articles are then summarized by GPT-4o to form concise Exogenous News, providing the necessary background context for the time series. Specific exogenous news summarization prompt is in Figure \ref{fig:exogenous text}.
\end{itemize}

\begin{figure}[hbp]
    \centering
    \includegraphics[width=0.9\linewidth]{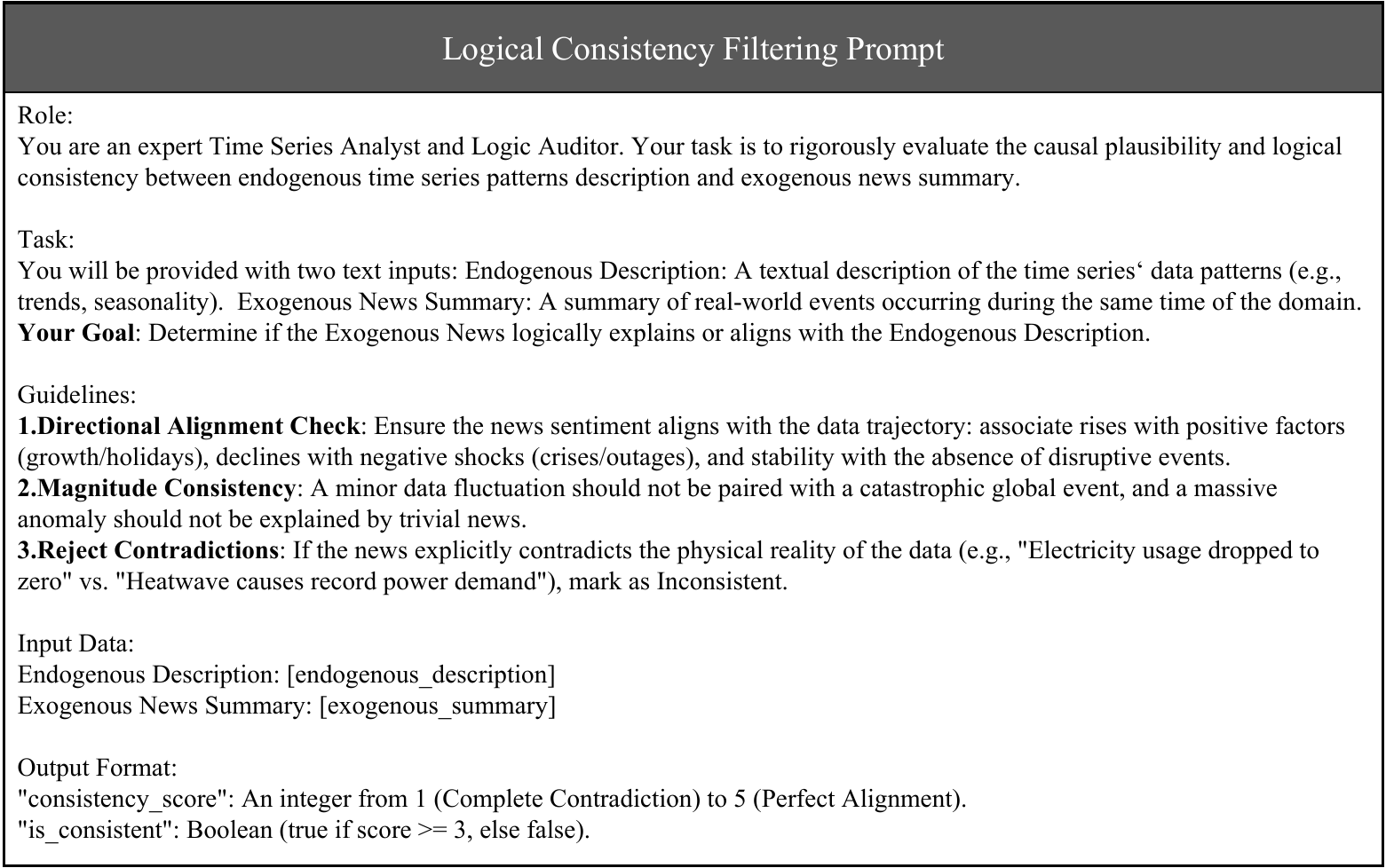}
    \caption{The prompt is designed for judgment logical consistency for the endogenous description and exogenous news.}
    \label{fig:logical consistency}
\end{figure}

Stage 2: Quality Alignment Stage. To ensure the reliability and logical consistency of the synthesized data, the raw generations undergo a rigorous quality control process:

\begin{itemize}
    \item 
Logical Consistency Check: A dedicated LLM GPT-4o evaluates the alignment between the Endogenous Text and Exogenous News. This step filters out hallucinations or irrelevant pairings by verifying, for example, that a significant market dip is explained by a corresponding negative news event rather than a positive one. Specific logical consistency filtering prompt is in Figure \ref{fig:logical consistency}.
    \item
Ensemble Quality Evaluation: To mitigate the bias inherent in single-model evaluation, the unified text is scrutinized by a heterogeneous ensemble of LLM judges. We employ a consensus-based filtering mechanism, where only samples achieving high aggregate scores on factual plausibility and semantic clarity are incorporated into the final dataset.

\end{itemize}

\subsection{Evaluation Dataset}
To evaluate HORAI in a multi-task setting, we employ widely used benchmark datasets for both forecasting and anomaly detection. 1) Forecasting: As shown in Table~\ref{tab:forecasting_datasets}, experiments are conducted on TimeMMD \citep{timemmd} and additional datasets \citep{FNSPID}, covering diverse domains such as Agriculture, Climate, Energy, Environment, Social Good, Traffic, EWJ, KR, and MDT. 2) Anomaly Detection: We evaluate HORAI on five datasets—Weather, Energy, KR, EWJ, and MDT—with anomaly ratios ranging from 5.81\% to 17.23\%. Detailed statistics are provided in Table~\ref{tab:anomaly_detection_datasets}.

\begin{table}[thbp]
  \caption{The statistics of evaluation datasets for the forecasting task.}
  \label{tab:forecasting_datasets}
  \centering
  \resizebox{0.8\linewidth}{!}{
  \setlength{\tabcolsep}{12pt}
  \begin{tabular}{l|l|c|c|c|c}
    \hline
    \hline
    \textbf{Tasks} & \textbf{Dataset} & \textbf{Variate} & \textbf{Frequency} & \textbf{Dataset Size} & \textbf{Timespan} \\
    \toprule
     & Agriculture & 1 & Monthly & 496 & 1983-2024\\
    \cmidrule{2-6}
     \multirow{9}{*}{Forecasting} & Climate & 5 & Monthly & 496 & 1983-2024 \\
    \cmidrule{2-6}
    & Energy & 9 & Weekly & 1479 & 1996-2024\\
    \cmidrule{2-6}
    & Environment & 4 & Daily & 11102 & 1982-2023 \\
    \cmidrule{2-6}
    & Social Good & 1 & Monthly & 900 & 1950-2024 \\
    \cmidrule{2-6}
    & Traffic & 1 & Monthly & 531 & 1980-2024 \\
    \cmidrule{2-6}
    & EWJ & 1 & Daily & 2658 & 2009-2020 \\
    \cmidrule{2-6}
    & KR & 1 & Daily & 2655 & 2009-2020 \\
    \cmidrule{2-6}
    & MDT & 1 & Daily & 2732 & 2009-2020 \\
    \hline
    \hline
    \end{tabular}
  }
\end{table}

\begin{table}[thbp]
  \caption{The statistics of evaluation datasets for the anomaly detection task.}
  \label{tab:anomaly_detection_datasets}
  \centering
  \resizebox{\linewidth}{!}{
  \setlength{\tabcolsep}{12pt}
  \begin{tabular}{l|l|c|c|c}
    \hline
    \hline
    \textbf{Tasks} & \textbf{Dataset} & \textbf{Anomaly Ratio} & \textbf{Frequency} & \textbf{Dataset Description} \\
    \toprule
     \multirow{5}{*}{Detection} & Weather & 17.10\% & Monthly & Temperature and humidity information collected from government websites. \\
    \cmidrule{2-5}
    & Energy & 17.23\% & Weekly & The dataset records weekly U.S. gasoline prices (dollars per gallon).
 \\
    \cmidrule{2-5}
    & KR & 6.21\% & Daily &  The dataset is collected from Yahoo, NASDAQ finance websites. \\
    \cmidrule{2-5}
    & MDT & 11.17\% & Monthly & The dataset is collected from Yahoo, NASDAQ finance websites. \\
    \cmidrule{2-5}
    & EWJ & 9.96\% & Daily &  The dataset is collected from Yahoo, NASDAQ finance websites.\\
    \hline
    \hline
    \end{tabular}
  }
\end{table}

\section{Baselines}
\label{appendix:baselines}

We categorize the baselines into three groups: \textit{Time Series Foundation Models}, \textit{Multimodal Time-Series-Specific Models}, and \textit{Unimodal Time-Series-Specific Models}. Time Series Foundation Models are pre-trained on large-scale, cross-domain unimodal time series data, enabling direct inference on downstream tasks and demonstrating generalization capabilities. In contrast, Time-Series-Specific Models require training on each downstream dataset and can be further divided by input type. Multimodal Time-Series-Specific Models leverage additional modalities, such as text or images, or reuse LLM representations to enhance time series understanding. Unimodal Time-Series-Specific Models, on the other hand, design tailored modules to exploit the inherent characteristics of time series data.

\subsection{Time Series Foundation Models}
\begin{itemize}
    \item ChatTime \citep{chattime} proposes a unified multimodal time series foundation model that tokenizes both numerical sequences and text prompts into a shared vocabulary, enabling forecasting and question answering tasks with a single generative language-model backbone. 
    \item Sundial \citep{sundial} proposes a TimeFlow Loss that predicts the distribution of the next patch, enabling Transformer training without discrete tokenization and supporting probabilistic forecasting.
    \item VisionTS \citep{visionts} converts time series data into image form and uses visual mask autoencoders for unsupervised feature learning.
    \item ROSE \citep{rose} combines frequency decomposition with time-series registers to jointly learn both domain-invariant and domain-specific representations, facilitating knowledge transfer to downstream tasks. 
    \item Timer \citep{timer} adopts a decoder-only architecture employing autoregressive modeling for generative pre-training.
    \item MOIRAI \citep{moirai} introduces multi-scale patch projections to model diverse patterns and an any-variate attention mechanism that allows flexible handling of time series with arbitrary dimensionality.
    \item DADA \citep{dada} leverages adaptive bottleneck and dual-adversarial decoding to enable robust zero-shot anomaly detection across diverse domains.
    \item UniTS \citep{units} proposes a novel unified network backbone for classification, forecasting, and anomaly detection.
\end{itemize}

\subsection{Multimodal Time-Series-Specific Models}
\begin{itemize}
    \item GPT4MTS \citep{gpt4mts} propose a prompt tuning-based LLM for time series forecasting with multimodal input.
    \item TATS \citep{tats} propose a plug-and-play multimodal time series forecasting
framework, which transforms text representations into auxiliary variables.
    \item GPT4TS \citep{gpt4ts} fine-tunes the limited parameters of LLM, demonstrating competitive performance by transferring knowledge from large-scale pre-training text data.
    \item LLMMixer \citep{llmmixer} adapts LLMs for time series forecasting by breaking down the data into different time scales.
    \item TimeVLM \citep{timevlm} leverages pre-trained VLMs to enhance time series forecasting by unifying temporal, visual, and textual information.
\end{itemize}

\subsection{Unimodal Time-Series-Sepcific Models}
\begin{itemize}
    \item TimesNet \citep{Timesnet} transforms the 1D time series into a set of 2D tensors based on multiple periods to handle the multi-periodicity of the time series.
    \item DCdetector \citep{DCdetector} leverages dual-attention contrastive representation learning, extracting normal feature representations through self-supervised learning and dual-attention mechanisms.
    \item Anomaly Transformer \citep{A-T} leverages a self-attention mechanism to capture both short- and long-term dependencies in time series, and detects anomalies by analyzing differences in association matrices.
    \item PatchTST \citep{patchtst} segments time series into subseries-level patches that serve as input tokens to the Transformer and applies the channel-independence strategy for training on multivariate time series.
    \item HBOS \citep{HBOS} is a fast unsupervised anomaly detection method based on histogram density estimation.
    \item IForest \citep{IForest} detects anomalies by recursively partitioning data to isolate outliers, rather than modeling normal behavior.
    \item PCA \citep{pca} detects anomalies by measuring deviations in the principal component space, assuming outliers lie far from the normal distribution.
\end{itemize}

\section{Experiment Setting}

\textit{During pre-training}, HORAI is optimized using the Adam optimizer with an initial learning rate of 0.0005 and trained for 20 epochs, with early stopping applied using a patience of 10 epochs. The batch size is set to 2048, the input time series length to 576, and the patch size to 48. The Time-Frequency Decoder is configured with 6 layers, the model dimension $D_{model}$ is set to 768, and the ratio parameter $\alpha$ for high- and low-frequency decomposition is fixed at 0.05. All experiments are implemented in PyTorch, and pre-training is conducted on four NVIDIA Tesla A800 80GB GPUs.

\textit{For forecasting}, 
To ensure fairness, we remove the drop-last strategy for HORAI and all baselines, since using it would result in inconsistent numbers of test samples across different batch sizes \cite{tfb}. For each dataset, we evaluate four prediction horizons for both HORAI and the baselines. Specifically, Agriculture, Climate, Social Good, Traffic, EWJ, KR, and MDT are evaluated with horizons $\{6, 8, 10, 12\}$, Environment with $\{48, 96, 192, 336\}$, and Energy with $\{12, 24, 36, 48\}$.

\section{Discussion with ChatTime}
\label{discussion chattime}
We further provide a clarified discussion comparing ChatTime and HORAI, including the following aspects: (1) \textbf{Model perspective}: HORAI is \textbf{specifically architected as a multimodal foundation model} integrating time series, images, and text. It leverages modality-specific encoders to extract distinct features and employs a novel frequency-enhanced alignment to fuse these representations from multiple perspectives explicitly. In contrast,  ChatTime \textbf{adapts general-purpose LLMs} for time series analysis. While leveraging LLMs' inherent reasoning abilities for time series analysis offers generalization, discretizing continuous numerical values into textual tokens results in precision loss, making it challenging to capture time series patterns.
(2) \textbf{Data perspective}: HORAI is pretrained on a large-scale multimodal dataset incorporating aligned text and images. These modalities capture diverse characterizations of temporal dynamics from multiple perspectives and simultaneously introduce some external context, providing relevant supervision that improves generalization. However, ChatTime relies only on simple prompts such as "Please predict the following sequence," which offer limited text regarding the specific time series characteristics.

\section{Model Analysis}

\subsection{Sensitivity analysis}
\label{sensitivity analysis}

We conduct sensitivity experiments on two key parameters: the frequency threshold $\alpha$ and the number of selected experts $K$. As shown in the Table \ref{tab:parameter frequency}, setting $\alpha$ to 0.05 achieves the best prediction performance. This value distinctly partitions low-frequency from mid-to-high-frequency features, facilitating optimal alignment with text and image modalities. Conversely, a larger $\alpha$ forces excessive information into high-frequency components, thereby amplifying noise-like patterns; whereas an overly small $\alpha$ introduces redundant low-frequency information, which disrupts the alignment between image and time series representations. As shown in Table \ref{tab:parameter the number of selected experts}, 
selecting the Top-2 or Top-3 experts yields superior performance. Activating all experts tends to introduce redundancy from irrelevant experts, thereby diluting the model's generalization. Whereas selecting only a single expert limits the representational capacity, preventing the model from modeling diverse patterns.

\begin{table}[h]
  \centering
  \caption{Hyper-parameter sensitivity analysis about the frequency threshold $\alpha$.}
    \begin{tabular}{ccccc}
    \hline
    \hline
          & \textbf{$\alpha$ = 0.01} & \textbf{$\alpha$= 0.05} & \textbf{$\alpha$ = 0.25} & \textbf{$\alpha$ = 0.5} \\
    \hline
    Metrics & MSE   & MSE   & MSE   & MSE \\
    \hline
    Agriculture & 0.210  & \textbf{0.201 } & 0.225  & 0.247  \\
    Climate & 0.868  & \textbf{0.857 } & 1.054  & 1.200  \\
    Energy & 0.228  & \textbf{0.208 } & 0.284  & 0.266  \\
    Environment & 0.315  & \textbf{0.309 } & 0.332  & 0.333  \\
    \hline
    \hline
    \end{tabular}%
  \label{tab:parameter frequency}%
\end{table}%

\begin{table}[h]
  \centering
  \caption{Hyper-parameter sensitivity analysis about the number of selected experts $K$.}
    \begin{tabular}{ccccc}
    \hline
    \hline
          & \textbf{K=1} & \textbf{K=2} & \textbf{K=3} & \textbf{K=4} \\
    \hline
    \multicolumn{1}{l}{Metrics} & MSE   & MSE   & MSE   & MSE \\
    \hline
    Agriculture & 0.227  & \textbf{0.201 } & 0.206  & 0.220  \\
    Climate & 1.048  & \textbf{0.857 } & 0.880  & 0.891  \\
    Energy & 0.217  & 0.208  & \textbf{0.200 } & 0.215  \\
    Environment & 0.321  & \textbf{0.309 } & 0.316  & 0.326  \\
    \hline
    \hline
    \end{tabular}%
  \label{tab:parameter the number of selected experts}%
\end{table}%

\subsection{Ablation Analysis on Specific Modalities and Alignment Strategies}
\label{specific modalities and alignment strategies}

We perform ablation studies to evaluate the contributions of individual modalities (text, image) and the efficacy of our frequency-based alignment strategy. Specifically, we analyze four settings: 1) only text and time series; 2) only image and time series ; 3) text, image, and time series without frequency-based alignment (w/o Freq-Align); and 4) swapping modalities by fusing low-frequency time series with images and mid-to-high frequency time series with text (Modality Exchange).
As shown in Table \ref{tab:different modalities}, both visual and textual modalities contribute to performance gains, though their relative impact varies depending on the dataset characteristics. For datasets exhibiting clear long-term trends, such as Agriculture and Energy, the text modality contributes more significantly. Conversely, for datasets dominated by local fluctuations, such as Climate, the image modality proves more critical. Crucially, the significant performance drop observed when removing frequency-based alignment and modality exchange underscores the validity of our design: it confirms that aligning images with mid-to-high frequency components and text with low-frequency components is the effective strategy.

\begin{table}[h]
  \centering
  \caption{Ablation analysis about each modality.}
  \resizebox{\linewidth}{!}{
    \begin{tabular}{ccccccccccc}
    \hline
    \hline 
          & \multicolumn{2}{c}{\textbf{HORAI}} & \multicolumn{2}{c}{\textbf{Text + Time Series}} & \multicolumn{2}{c}{\textbf{Image + Time Series}} & \multicolumn{2}{c}{\textbf{W/O Freq-Align}} & \multicolumn{2}{c}{\textbf{Modality Exchange}} \\
    \hline 
    Metrics & MSE   & MAE   & MSE   & MAE   & MSE   & MAE   & MSE   & MAE   & MSE   & MAE \\
    \hline 
     Agriculture & \textbf{0.201 } & \textbf{0.309 } & 0.248  & 0.327  & 0.275  & 0.340  & 0.234  & 0.324  & 0.292  & 0.352  \\
    Climate & \textbf{0.857 } & \textbf{0.734 } & 1.102  & 0.828  & 0.982  & 0.797  & 0.928  & 0.786  & 1.321  & 0.856  \\
    Energy & \textbf{0.208 } & \textbf{0.325 } & 0.255  & 0.372  & 0.276  & 0.395  & 0.248  & 0.355  & 0.292  & 0.402  \\
    Environment & \textbf{0.309 } & \textbf{0.393 } & 0.344  & 0.412  & 0.320  & 0.398  & 0.325  & 0.396  & 0.360  & 0.426  \\
    \hline 
    \hline 
    \end{tabular}%
    }
  \label{tab:different modalities}%
\end{table}%

\subsection{Ablation Analysis about Text Encoder and Vision Encoder}
\label{text encoder and vision encoder}
To evaluate the model’s performance with different encoders, we conduct additional experiments by replacing both text and visual encoders. By default, HORAI employs Qwen-0.5B and ViT-Base for text and visual modalities, respectively. Considering time and computational constraints, we select encoders with relatively small parameter sizes. Specifically, the text encoders include GPT2-small, LLaMA3-1B, and Qwen2.5-1.5B, while the visual encoder comparison uses Swin Transformer.  As shown in the Table \ref{tab:different encoders}, for a given text encoder, models with larger parameter sizes tend to perform slightly better, and employing more advanced architectures (e.g., Qwen and LLaMA) generally yields further improvements. In the comparison of visual encoders, ViT and Swin Transformer achieve similar overall forecasting performance.  
\begin{table}[h]
  \centering
  \caption{Ablation analysis of different text encoders and image encoders.}
   \resizebox{\linewidth}{!}{
    \begin{tabular}{ccccccccccc}
    \hline
    \hline 
          & \multicolumn{2}{c}{\textbf{HORAI}} & \multicolumn{2}{c}{\textbf{GPT2}} & \multicolumn{2}{c}{\textbf{Llama3-1B}} & \multicolumn{2}{c}{\textbf{Qwen-1.5B}} & \multicolumn{2}{c}{\textbf{Swin Trans-Base}} \\
    \hline 
    Metrics & MSE   & MAE   & MSE   & MAE   & MSE   & MAE   & MSE   & MAE   & MSE   & MAE \\
    \hline
     Agriculture & 0.201  & 0.309  & 0.238  & 0.325  & 0.210  & 0.315  & 0.204  & 0.314  & \textbf{0.198 } & \textbf{0.302 } \\
    Climate & 0.857  & 0.734  & 0.913  & 0.840  & 0.851  & 0.730  & \textbf{0.832 } & \textbf{0.715 } & 0.863  & 0.758  \\
    Energy & 0.208  & 0.325  & 0.229  & 0.348  & 0.204  & 0.322  & \textbf{0.202 } & \textbf{0.318 } & 0.212  & 0.336  \\
    Environment & 0.309  & 0.393  & 0.325  & 0.398  & 0.310  & 0.395  & 0.302  & 0.390  & \textbf{0.300 } & \textbf{0.388 } \\
    \hline
    \hline
    \end{tabular}%
    }
  \label{tab:different encoders}%
\end{table}%

\begin{table}[htbp]
  \centering
  \caption{Full time series forecasting results of HORAI, time series foundation models, and time-series-specific models.}
   \renewcommand{\arraystretch}{1.5}
  \resizebox{\linewidth}{!}{
    \begin{tabular}{c|c|cccccccccccc|cccccccc}
    \hline
    \hline 
    \multicolumn{2}{c|}{\textbf{Type}} & \multicolumn{12}{c|}{\emoji{Figures/prohibit.png} \textbf{Time Series Foundation Models (Zero-Shot)}}                                    & \multicolumn{8}{c}{\emoji{Figures/deadline.png} \textbf{Time-Series-Specific-Models (Full-Shot)}} \\
    \hline
      \multicolumn{2}{c|}{\textbf{Models}} & \multicolumn{2}{c}{\textbf{HORAI}} & \multicolumn{2}{c}{\textbf{ChatTime}} & \multicolumn{2}{c}{\textbf{VisionTS}} & \multicolumn{2}{c}{\textbf{ROSE}} & \multicolumn{2}{c}{\textbf{Timer}} & \multicolumn{2}{c|}{\textbf{MOIRAI}} & \multicolumn{2}{c}{\textbf{GPT4MTS}} & \multicolumn{2}{c}{\textbf{TATS}} & \multicolumn{2}{c}{\textbf{GPT4TS}} & \multicolumn{2}{c}{\textbf{TimeVLM}} \\
      \hline
    \multicolumn{2}{c|}{\textbf{Metric}} & MSE   & MAE   & MSE   & MAE   & MSE   & MAE   & MSE   & MAE   & MSE   & MAE   & MSE   & MAE   & MSE   & MAE   & MSE   & MAE   & MSE   & MAE   & MSE   & MAE \\
    \hline 
     \multirow{5}[0]{*}{\textbf{Agriculture}} & 6     & \textbf{0.131 } & 0.260  & 0.243  & 0.340  & 0.210  & 0.289  & 0.219  & 0.299  & 0.168  & 0.272  & 0.187  & 0.342  & 0.161  & 0.257  & 0.140  & 0.251  & \underline{0.135 } & \textbf{0.242 } & 0.143  & \underline{0.245 } \\
          & 8     & \textbf{0.171 } & 0.294  & 0.349  & 0.399  & 0.266  & 0.323  & 0.278  & 0.339  & 0.243  & 0.317  & 0.245  & 0.391  & 0.207  & 0.288  & \underline{0.187 } & \textbf{0.282 } & 0.198  & \underline{0.284 } & 0.215  & 0.287  \\
          & 10    & \textbf{0.219 } & 0.323  & 0.390  & 0.418  & 0.307  & 0.348  & 0.408  & 0.406  & 0.328  & 0.361  & 0.297  & 0.423  & 0.230  & 0.305  & \underline{0.244 } & \underline{0.320 } & 0.258  & \textbf{0.313 } & 0.271  & 0.320  \\
          & 12    & \textbf{0.286 } & 0.362  & 0.497  & 0.483  & 0.376  & 0.386  & 0.474  & 0.443  & 0.415  & 0.405  & 0.357  & 0.455  & 0.301  & \underline{0.342 } & \underline{0.290 } & 0.350  & 0.291  & \textbf{0.338 } & 0.322  & 0.359  \\
          & avg   & \textbf{0.201 } & 0.309  & 0.369  & 0.410  & 0.290  & 0.336  & 0.345  & 0.372  & 0.289  & 0.339  & 0.272  & 0.403  & 0.225  & \underline{0.298 } & \underline{0.215 } & 0.301  & 0.220  & \textbf{0.294 } & 0.237  & 0.302  \\
    \hline
    \multirow{5}[0]{*}{\textbf{Climate}} & 6     & \textbf{0.854 } & \textbf{0.736 } & 1.884  & 1.118  & 1.316  & 0.932  & 1.488  & 0.993  & \underline{0.876 } & \underline{0.759 } & 1.624  & 1.016  & 1.199  & 0.895  & 1.194  & 0.897  & 1.207  & 0.901  & 1.218  & 0.907  \\
          & 8     & \textbf{0.852 } & \textbf{0.733 } & 1.843  & 1.100  & 1.312  & 0.935  & 1.598  & 1.031  & \underline{0.885 } & \underline{0.763 } & 2.148  & 1.152  & 1.205  & 0.899  & 1.178  & 0.886  & 1.191  & 0.892  & 1.181  & 0.914  \\
          & 10    & \textbf{0.859 } & \textbf{0.734 } & 1.806  & 1.090  & 1.302  & 0.928  & 1.401  & 0.967  & \underline{0.893 } & \underline{0.766 } & 1.983  & 1.112  & 1.173  & 0.885  & 1.170  & 0.881  & 1.169  & 0.886  & 1.179  & 0.880  \\
          & 12    & \textbf{0.865 } & \textbf{0.736 } & 1.909  & 1.117  & 1.297  & 0.925  & 1.414  & 0.957  & \underline{0.899 } & \underline{0.770 } & 1.929  & 1.101  & 1.152  & 0.876  & 1.179  & 0.885  & 1.171  & 0.883  & 1.203  & 0.896  \\
          & avg   & \textbf{0.857 } & \textbf{0.734 } & 1.860  & 1.106  & 1.307  & 0.930  & 1.475  & 0.987  & \underline{0.888 } & \underline{0.764 } & 1.921  & 1.095  & 1.182  & 0.889  & 1.180  & 0.887  & 1.184  & 0.891  & 1.195  & 0.899  \\
    \hline
    \multirow{5}[0]{*}{\textbf{Energy}} & 12    & \underline{0.105 } & \underline{0.226 } & \textbf{0.104 } & \textbf{0.222 } & 0.173  & 0.313  & 0.268  & 0.401  & 0.118  & 0.236  & 0.183  & 0.309  & 0.111  & 0.244  & 0.105  & 0.232  & 0.111  & 0.243  & 0.114  & 0.253  \\
          & 24    & \textbf{0.189 } & \textbf{0.314 } & \underline{0.203 } & \underline{0.321 } & 0.264  & 0.395  & 0.363  & 0.469  & 0.225  & 0.336  & 0.290  & 0.396  & 0.232  & 0.362  & 0.216  & 0.344  & 0.223  & 0.355  & 0.227  & 0.359  \\
          & 36    & \textbf{0.248 } & \textbf{0.362 } & \underline{0.292 } & \underline{0.396 } & 0.346  & 0.454  & 0.413  & 0.497  & 0.328  & 0.403  & 0.367  & 0.449  & 0.308  & 0.418  & 0.309  & 0.418  & 0.314  & 0.423  & 0.309  & 0.410  \\
          & 48    & \textbf{0.292 } & \textbf{0.399 } & 0.389  & 0.470  & 0.434  & 0.516  & 0.501  & 0.549  & 0.424  & \underline{0.460 } & 0.457  & 0.515  & 0.398  & 0.496  & \underline{0.391 } & 0.480  & 0.393  & 0.484  & 0.390  & 0.475  \\
          & avg   & \textbf{0.208 } & \textbf{0.325 } & \underline{0.247 } & \underline{0.352 } & 0.304  & 0.420  & 0.386  & 0.479  & 0.274  & 0.359  & 0.324  & 0.417  & 0.262  & 0.380  & 0.255  & 0.368  & 0.260  & 0.376  & 0.260  & 0.374  \\
    \hline
    \multirow{5}[0]{*}{\textbf{Environment}} & 48    & \underline{0.306 } & \textbf{0.379 } & 0.343  & 0.406  & 0.345  & 0.426  & 0.402  & 0.459  & 0.358  & 0.431  & 0.352  & 0.404  & 0.315  & 0.400  & 0.307  & 0.389  & 0.320  & 0.396  & \textbf{0.304 } & \underline{0.387 } \\
          & 96    & \textbf{0.326 } & \textbf{0.396 } & 0.369  & 0.465  & 0.370  & 0.441  & 0.409  & 0.465  & 0.368  & 0.436  & 0.370  & 0.415  & 0.340  & 0.401  & 0.334  & 0.402  & 0.340  & 0.401  & \underline{0.327 } & \underline{0.405 } \\
          & 192   & \textbf{0.306 } & \textbf{0.395 } & 0.377  & 0.474  & 0.360  & 0.442  & 0.389  & 0.452  & 0.351  & 0.427  & 0.350  & 0.402  & 0.336  & 0.411  & 0.332  & \underline{0.401 } & 0.330  & 0.391  & \underline{0.328 } & 0.403  \\
          & 336   & \textbf{0.298 } & \underline{0.403 } & 0.372  & 0.478  & 0.340  & 0.436  & 0.369  & 0.447  & 0.326  & 0.418  & 0.332  & 0.390  & 0.299  & 0.390  & 0.302  & 0.391  & \underline{0.300 } & \textbf{0.383 } & 0.320  & 0.395  \\
          & avg   & \textbf{0.309 } & \textbf{0.393 } & 0.359  & 0.456  & 0.354  & 0.436  & 0.392  & 0.456  & 0.351  & 0.428  & 0.351  & 0.403  & 0.323  & 0.400  & 0.319  & 0.396  & 0.322  & \underline{0.393 } & \underline{0.319 } & 0.397  \\
    \hline
    \multirow{5}[0]{*}{\textbf{Social Good}} & 6     & \textbf{0.682 } & 0.418  & 0.988  & 0.451  & 0.957  & 0.543  & 0.939  & 0.499  & 0.845  & 0.416  & 0.966  & 0.522  & 0.718  & 0.382  & 0.753  & \textbf{0.370 } & \underline{0.717 } & \underline{0.374 } & 0.732  & 0.379  \\
          & 8     & \textbf{0.779 } & 0.474  & 1.044  & 0.488  & 1.106  & 0.605  & 1.168  & 0.588  & 0.938  & 0.469  & 1.532  & 0.653  & 0.942  & 0.505  & 0.875  & \textbf{0.409 } & \underline{0.855 } & 0.459  & 0.822  & \underline{0.427 } \\
          & 10    & \textbf{0.869 } & 0.528  & 1.098  & 0.519  & 1.164  & 0.636  & 1.187  & 0.595  & 1.018  & 0.515  & 1.551  & 0.691  & 0.929  & \textbf{0.446 } & 0.991  & \underline{0.459 } & 0.930  & 0.463  & \underline{0.916 } & 0.465  \\
          & 12    & \textbf{0.948 } & 0.569  & 1.149  & 0.554  & 1.278  & 0.688  & 1.272  & 0.642  & 1.094  & 0.557  & 1.671  & 0.736  & 1.093  & \textbf{0.470 } & 1.053  & \underline{0.474 } & 1.167  & 0.608  & \underline{1.005 } & 0.505  \\
          & avg   & \textbf{0.819 } & 0.497  & 1.069  & 0.503  & 1.126  & 0.618  & 1.141  & 0.581  & 0.974  & 0.489  & 1.430  & 0.651  & 0.920  & 0.451  & 0.918  & \textbf{0.428 } & 0.917  & 0.476  & 0.868  & 0.444  \\
    \hline
    \multirow{5}[0]{*}{\textbf{Traffic}} & 6     & \textbf{0.163 } & \underline{0.240 } & 0.609  & 0.623  & 0.275  & 0.411  & 0.331  & 0.449  & 0.167  & 0.267  & 0.349  & 0.448  & 0.192  & 0.264  & \underline{0.164 } & \textbf{0.226 } & 0.199  & 0.278  & 0.210  & 0.316  \\
          & 8     & \underline{0.166 } & \underline{0.245 } & 0.626  & 0.636  & 0.282  & 0.410  & 0.365  & 0.455  & 0.185  & 0.287  & 0.461  & 0.499  & 0.195  & 0.256  & \textbf{0.178 } & \textbf{0.242 } & 0.204  & 0.262  & 0.212  & 0.313  \\
          & 10    & \textbf{0.164 } & \underline{0.246 } & 0.572  & 0.592  & 0.286  & 0.406  & 0.326  & 0.443  & 0.196  & 0.299  & 0.414  & 0.466  & 0.204  & 0.257  & \underline{0.185 } & \textbf{0.243 } & 0.210  & 0.264  & 0.222  & 0.328  \\
          & 12    & \textbf{0.165 } & \underline{0.247 } & 0.579  & 0.592  & 0.282  & 0.402  & 0.342  & 0.458  & 0.202  & 0.307  & 0.400  & 0.458  & 0.218  & 0.268  & \underline{0.189 } & \textbf{0.242 } & 0.211  & 0.260  & 0.222  & 0.322  \\
          & avg   & \textbf{0.165 } & \underline{0.244 } & 0.596  & 0.610  & 0.281  & 0.407  & 0.341  & 0.451  & 0.188  & 0.290  & 0.406  & 0.468  & 0.203  & 0.261  & \underline{0.179 } & \textbf{0.238 } & 0.206  & 0.266  & 0.216  & 0.319  \\
    \hline
    \multirow{5}[0]{*}{\textbf{EWJ}} & 6     & \textbf{0.548 } & 0.528  & 0.808  & 0.612  & 0.583  & 0.560  & 0.634  & 0.581  & 0.643  & 0.573  & 0.751  & 0.623  & 0.579  & 0.531  & 0.550  & 0.525  & \underline{0.550 } & \underline{0.523 } & 0.552  & \textbf{0.521 } \\
          & 8     & \textbf{0.581 } & \textbf{0.537 } & 0.880  & 0.641  & 0.629  & 0.580  & 0.729  & 0.626  & 0.685  & 0.591  & 1.017  & 0.714  & 0.608  & 0.540  & 0.611  & 0.544  & \underline{0.597 } & \underline{0.538 } & 0.599  & 0.541  \\
          & 10    & \textbf{0.604 } & \textbf{0.550 } & 0.920  & 0.652  & 0.665  & 0.591  & 0.716  & 0.599  & 0.716  & 0.604  & 0.982  & 0.705  & 0.644  & 0.559  & \underline{0.627 } & \textbf{0.551 } & 0.632  & 0.551  & 0.629  & 0.554  \\
          & 12    & \textbf{0.623 } & \textbf{0.558 } & 0.940  & 0.659  & 0.701  & 0.607  & 0.746  & 0.613  & 0.740  & 0.614  & 0.997  & 0.709  & 0.673  & 0.566  & 0.661  & 0.563  & \underline{0.649 } & \underline{0.560 } & 0.657  & 0.562  \\
          & avg   & \textbf{0.589 } & \textbf{0.543 } & 0.887  & 0.641  & 0.645  & 0.584  & 0.706  & 0.605  & 0.696  & 0.595  & 0.937  & 0.688  & 0.626  & 0.549  & 0.612  & 0.546  & \underline{0.607 } & \textbf{0.543 } & 0.609  & 0.544  \\
    \hline
    \multirow{5}[0]{*}{\textbf{KR}} & 6     & \textbf{0.533 } & \textbf{0.435 } & 0.528  & 0.436  & 0.628  & 0.503  & 0.687  & 0.521  & 0.530  & 0.453  & 0.793  & 0.567  & 0.528  & 0.442  & 0.542  & \underline{0.426 } & \underline{0.539 } & 0.435  & 0.550  & 0.437  \\
          & 8     & \textbf{0.549 } & \textbf{0.446 } & 0.564  & 0.452  & 0.674  & 0.524  & 0.798  & 0.572  & 0.547  & 0.461  & 1.077  & 0.650  & 0.564  & \underline{0.452 } & 0.569  & 0.446  & 0.573  & 0.444  & 0.580  & 0.451  \\
          & 10    & 0.561  & \textbf{0.454 } & 0.570  & 0.459  & 0.685  & 0.526  & 0.727  & 0.530  & \textbf{0.559 } & 0.468  & 1.063  & 0.649  & 0.566  & 0.455  & 0.600  & 0.462  & 0.594  & \underline{0.452 } & 0.601  & 0.463  \\
          & 12    & \textbf{0.562 } & \underline{0.458 } & 0.598  & 0.473  & 0.698  & 0.535  & 0.750  & 0.547  & 0.560  & 0.472  & 1.038  & 0.649  & \underline{0.562 } & \textbf{0.453 } & 0.602  & 0.461  & 0.604  & 0.459  & 0.606  & 0.467  \\
          & avg   & \textbf{0.551 } & \textbf{0.448 } & 0.565  & 0.455  & 0.671  & 0.522  & 0.741  & 0.542  & 0.549  & 0.463  & 0.992  & 0.629  & \underline{0.555 } & \underline{0.450 } & 0.578  & 0.449  & 0.578  & 0.448  & 0.584  & 0.454  \\
    \hline
    \multirow{5}[0]{*}{\textbf{MDT}} & 6     & \textbf{0.361 } & 0.426  & 0.466  & 0.455  & 0.412  & 0.471  & 0.426  & 0.476  & 0.366  & 0.437  & 0.494  & 0.521  & \underline{0.369 } & 0.436  & 0.365  & \underline{0.423 } & 0.373  & \textbf{0.422 } & 0.369  & \underline{0.423 } \\
          & 8     & \textbf{0.372 } & \textbf{0.433 } & 0.474  & 0.473  & 0.431  & 0.486  & 0.483  & 0.514  & 0.383  & 0.446  & 0.668  & 0.591  & \underline{0.377 } & \underline{0.439 } & 0.383  & 0.433  & 0.386  & 0.432  & 0.385  & 0.434  \\
          & 10    & \textbf{0.382 } & \textbf{0.438 } & 0.526  & 0.494  & 0.437  & 0.487  & 0.456  & 0.486  & 0.397  & 0.453  & 0.630  & 0.580  & \underline{0.389 } & 0.444  & 0.397  & \underline{0.440 } & 0.395  & 0.448  & 0.400  & 0.443  \\
          & 12    & \textbf{0.390 } & \textbf{0.442 } & 0.518  & 0.494  & 0.453  & 0.495  & 0.477  & 0.499  & 0.408  & 0.458  & 0.632  & 0.582  & \underline{0.405 } & 0.450  & 0.411  & \underline{0.447 } & 0.411  & 0.452  & 0.414  & 0.448  \\
          & avg   & \textbf{0.376 } & \textbf{0.434 } & 0.496  & 0.479  & 0.433  & 0.485  & 0.461  & 0.493  & 0.389  & 0.448  & 0.606  & 0.569  & \underline{0.385 } & 0.442  & 0.389  & \underline{0.436 } & 0.391  & 0.438  & 0.392  & 0.437  \\
    \hline
    \rowc
    \multicolumn{2}{c}{\textbf{1\textsuperscript{st} Count}} & \textbf{\textbf{41}} & \textbf{25} & 1 & 1 & 0 & 0 & 0 & 0 & 1 & 0 & 0 & 0 & 0 & 3 & 1 & 10 & 0 & 7 & 1 & 1  \\
    \hline
    \hline
    \end{tabular}%
  }
  \label{tab:forecasting_full}%
\end{table}%

\begin{table}[htbp]
  \centering
  \caption{Time series anomaly detection results under zero-shot and full-shot settings with multiple metrics. The best results are in \textbf{bold}, and the second-best results are \underline{underlined}.}
   \resizebox{\linewidth}{!}{
    \begin{tabular}{c|c|cccc|ccccccccc}
    \hline
    \hline
    \multirow{2}[2]{*}{\textbf{Type}} & \multicolumn{5}{c|}{\multirow{2}[2]{*}{\emoji{Figures/prohibit.png} \textbf{Time Series Foundation Models (Zero-Shot)}}} & \multicolumn{9}{c}{\multirow{2}[2]{*}{\emoji{Figures/deadline.png} \textbf{Time-Series-Specific-Models (Full-Shot)}}} \\
          & \multicolumn{5}{c|}{}                 & \multicolumn{9}{c}{} \\
    \midrule
    \textbf{Datasets} & \textbf{Metric} & \textbf{HORAI} & \textbf{DADA} & \textbf{Timer} & \textbf{UniTS} & \textbf{GPT4TS} & \textbf{LLMMixer} & \textbf{TimesNet} & \textbf{DCdetector} & \textbf{A.T.} & \textbf{PatchTST} & \textbf{HBOS} & \textbf{IForest} & \textbf{PCA} \\
    \midrule
    \multirow{8}[16]{*}{\textbf{EWJ}} & Aff-F1 & \textbf{82.54} & 81.26 & 78.06 & 77.61 & 76.65 & 66.86  & \underline{81.82} & 48.10 & 59.03 & 75.82 & 71.03 & 67.55 & 51.06 \\
\cmidrule{2-15}          & F1    & \textbf{56.28} & 49.33 & 41.21 & 39.18 & 48.33 & 18.95  & \underline{49.37} & 17.09 & 14.39 & 45.39 & 44.80 & 41.67 & 18.68 \\
\cmidrule{2-15}          & Range-AUC-ROC & \textbf{79.84} & 68.44 & 64.71 & 67.15 & 55.57 & 41.75 & \underline{74.22} & 45.69 & 27.50 & 69.26 & 61.02 & 57.94 & 43.78 \\
\cmidrule{2-15}          & Range-AUC-PR & \textbf{43.33} & 41.61 & 30.67 & 44.31 & 32.83 & 12.36  & \underline{41.84} & 12.52 & 9.01  & 33.37 & 46.37 & 34.77 & 16.49 \\
\cmidrule{2-15}          & AUC-PR & 51.97 & \textbf{55.24} & 44.01 & 50.33 & 46.75 & 18.81  & \underline{54.99} & 10.88 & 8.97  & 47.91 & 25.24 & 24.16 & 10.99 \\
\cmidrule{2-15}          & AUC-ROC & \textbf{86.95} & 79.11 & 76.15 & 79.87 & 75.58 & 57.69  & \underline{82.39} & 53.40 & 43.81 & 78.53 & 71.82 & 69.20 & 54.35 \\
\cmidrule{2-15}          & VUS-ROC & \textbf{82.78} & 71.79 & 67.72 & 73.91 & 67.95 & 52.79  & \underline{75.76} & 47.10 & 31.75 & 71.96 & 62.07 & 59.24 & 45.26 \\
\cmidrule{2-15}          & VUS-PR & \textbf{48.27} & 43.36 & 33.17 & 39.32 & 35.63 & 15.13  & \underline{43.15} & 15.37 & 10.85 & 36.08 & 41.19 & 37.81 & 19.38 \\
    \midrule
    \multirow{8}[16]{*}{\textbf{MDT}} & Aff-F1 & \textbf{80.66} & 77.99 & 78.51 & 75.57 & 80.81 & 67.65  & \underline{80.08} & 47.33 & 66.12 & 79.47 & 52.33 & 53.74 & 54.66 \\
\cmidrule{2-15}          & F1    & \textbf{59.36} & 53.70 & 48.39 & 51.70 & 49.14 & 27.71  & \underline{54.88} & 19.54 & 25.46 & 49.40 & 43.84 & 38.10 & 20.75 \\
\cmidrule{2-15}          & Range-AUC-ROC & \textbf{86.59} & 63.94 & 58.98 & 58.78 & 59.00 & 42.44  & \underline{77.01} & 43.65 & 41.41 & 75.61 & 54.86 & 53.22 & 41.90 \\
\cmidrule{2-15}          & Range-AUC-PR & \textbf{51.11} & 44.63 & 37.54 & 36.14 & 42.48 & 15.30  & \underline{48.60} & 13.30 & 13.20 & 13.11 & 43.16 & 33.63 & 19.53 \\
\cmidrule{2-15}          & AUC-PR & 61.98 & \underline{63.03} & 55.86 & 53.44 & 60.40 & 19.86  & \textbf{65.57} & 11.59 & 15.29 & 54.11 & 28.66 & 22.41 & 12.29 \\
\cmidrule{2-15}          & AUC-ROC & \textbf{91.22} & 79.04 & 75.65 & 73.19 & 74.79 & 60.30  & \underline{86.67} & 53.82 & 56.44 & 84.55 & 60.26 & 63.92 & 54.51 \\
\cmidrule{2-15}          & VUS-ROC & \textbf{86.82} & 66.76 & 60.28 & 58.67 & 62.30 & 46.80  & \underline{83.40} & 45.02 & 44.53 & 77.69 & 55.30 & 54.02 & 44.09 \\
\cmidrule{2-15}          & VUS-PR & \textbf{56.88} & 46.81 & 38.38 & 37.61 & 44.81 & 15.21  & \underline{52.13} & 15.72 & 15.93 & 41.67 & 44.77 & 35.32 & 22.93 \\
    \midrule
    \multirow{8}[16]{*}{\textbf{KR}} & Aff-F1 & \textbf{85.44} & 84.22 & 89.55 & 82.24 & 79.56 & 71.80 & 85.47 & 61.94 & 70.99 & 79.52 & 64.78 & 69.38 & 58.11 \\
\cmidrule{2-15}          & F1    & \textbf{71.89} & 49.48 & 58.04 & 30.23 & 74.01 & 20.25 & 58.14 & 11.98 & 11.10 & 36.64 & \underline{60.71} & 53.97 & 22.76 \\
\cmidrule{2-15}          & Range-AUC-ROC & \textbf{86.16} & 69.91 & 74.61 & 71.29 & 65.15 & 49.01 & \underline{78.29} & 41.75 & 40.18 & 72.72 & 61.80 & 61.10 & 51.01 \\
\cmidrule{2-15}          & Range-AUC-PR & \textbf{59.64} & 46.95 & 51.59 & \textbf{40.75} & 37.53 & 13.25 & \underline{52.83} & 6.04  & 7.44  & 35.22 & 51.69 & 43.07 & 18.99 \\
\cmidrule{2-15}          & AUC-PR & \textbf{72.91} & 63.55 & 66.72 & 55.39 & 56.78 & 28.19 & \underline{67.47} & 8.10  & 7.01  & 53.60 & 41.09 & 32.21 & 10.18 \\
\cmidrule{2-15}          & AUC-ROC & \textbf{96.38} & 79.53 & 66.72 & 80.95 & 78.30 & 65.77  & \underline{85.88} & 52.97 & 51.25 & 82.15 & 75.16 & 74.45 & 63.58 \\
\cmidrule{2-15}          & VUS-ROC & \textbf{93.54} & 70.82 & 75.99 & 73.93 & 67.81 & 47.06  & \underline{79.00} & 43.04 & 41.97 & 74.65 & 58.77 & 60.70 & 47.51 \\
\cmidrule{2-15}          & VUS-PR & \textbf{60.76} & 45.90 & 51.41 & 43.32 & 38.23 & 19.10  & \underline{51.60} & 8.49  & 7.94  & 36.18 & 54.17 & 43.31 & 24.19 \\
    \midrule
    \multirow{8}[16]{*}{\textbf{Energy}} & Aff-F1 & \textbf{71.37} & 64.38 & 60.20 & 63.84 & \underline{66.37} & 65.85 & 66.00 & 47.07 & 43.39 & 66.85 & 55.85 & 62.03 & 57.65 \\
\cmidrule{2-15}          & F1    & \textbf{37.71} & 31.54 & 31.71 & 31.66 & 33.22 & 33.08 & 33.95 & 12.63 & 12.05 & 34.81 & \underline{34.83} & 34.39 & 35.12 \\
\cmidrule{2-15}          & Range-AUC-ROC & \textbf{62.93} & 55.78 & 46.82 & 52.12 & 53.54 & 55.25 & \underline{61.56} & 45.39 & 31.52 & 61.39 & 51.06 & 52.64 & 52.64 \\
\cmidrule{2-15}          & Range-AUC-PR & 33.24 & 33.47 & 28.81 & 30.70 & 31.10 & 30.59 & \underline{38.17} & 21.77 & 19.24 & 35.25 & 42.14 & 45.19 & \textbf{43.89} \\
\cmidrule{2-15}          & AUC-PR & \underline{39.82} & 37.81 & 38.05 & 27.51 & 33.75 & 32.85 & \textbf{42.05} & 17.69 & 14.02 & 34.25 & 21.55 & 21.17 & 21.69 \\
\cmidrule{2-15}          & AUC-ROC & \textbf{68.44} & 62.33 & 60.54 & 63.38 & 66.54 & 61.31  & \underline{68.36} & 48.75 & 38.68 & 66.70 & 60.80 & 60.32 & 61.14 \\
\cmidrule{2-15}          & VUS-ROC & \textbf{62.42} & 54.37 & 46.03 & 51.15 & 53.10 & 53.04  & \underline{59.47} & 45.93 & 31.56 & 58.31 & 51.50 & 53.61 & 53.07 \\
\cmidrule{2-15}          & VUS-PR & 35.24 & 34.18 & 29.46 & 31.04 & 31.68 & 30.35  & 38.61 & 22.57 & 19.69 & 34.41 & \underline{42.57} & \textbf{46.03} & 44.30 \\
    \midrule
    \multirow{8}[16]{*}{\textbf{Weather}} & Aff-F1 & \textbf{80.84} & 69.01 & 75.46 & 76.17 & 72.56 & 73.68 & \underline{80.58} & 42.80 & 49.22 & 77.17 & 47.70 & 54.06 & 64.91 \\
\cmidrule{2-15}          & F1    & 47.44 & 35.29 & 46.42 & \underline{50.00} & 40.16 & 43.13 & \textbf{51.58} & 11.14 & 15.59 & 49.60 & 42.94 & 49.21 & 40.41 \\
\cmidrule{2-15}          & Range-AUC-ROC & \underline{80.61} & 61.95 & 73.37 & 75.55 & 71.43 & 72.54 & \textbf{83.11} & 45.41 & 43.11 & 80.47 & 54.12 & 56.69 & 57.80 \\
\cmidrule{2-15}          & Range-AUC-PR & \textbf{50.88} & 29.86 & 43.20 & 44.31 & 41.37 & 43.40 & \underline{50.58} & 18.06 & 18.85 & 49.81 & 46.37 & 49.65 & 47.47 \\
\cmidrule{2-15}          & AUC-PR & 49.16 & 29.80 & 48.87 & \underline{49.91} & 44.12 & 49.71 & 47.56 & 17.08 & 16.71 & \textbf{53.39} & 31.16 & 35.44 & 25.02 \\
\cmidrule{2-15}          & AUC-ROC & \textbf{81.49} & 66.37 & 80.86 & 81.22 & 74.47 & 79.60  & \underline{81.10} & 47.90 & 47.11 & 82.02 & 64.47 & 67.81 & 67.71 \\
\cmidrule{2-15}          & VUS-ROC & \underline{80.40} & 61.03 & 73.22 & 75.08 & 70.03 & 71.71  & \textbf{81.91} & 45.56 & 43.32 & 79.97 & 54.16 & 56.45 & 57.38 \\
\cmidrule{2-15}          & VUS-PR & \textbf{50.76} & 30.00 & 43.21 & 44.35 & 41.30 & 43.47  & 50.09 & 18.33 & 19.17 & \underline{50.13} & 46.58 & 49.66 & 47.13 \\
    \rowc
    \multicolumn{2}{c}{\textbf{1\textsuperscript{st} Count}} & \textbf{\textbf{31}} & 1 & 0 & 0 & 0 & 0 & 5 & 0 & 0 & 1 & 0 & 1 & 1  \\
    \hline
    \hline 
    \end{tabular}%
    }
  \label{tab: anomaly detection results with multiple metrics}%
\end{table}%


\end{document}